\documentclass[journal]{IEEEtran}
\usepackage[ruled,linesnumbered]{algorithm2e}
\usepackage{amsmath,epsfig,amssymb}
\usepackage[american]{babel}
\usepackage{booktabs}
\usepackage{hyperref}
\usepackage{enumitem}
\usepackage{array}
\usepackage{tabularx}
\usepackage{caption}
\usepackage{lipsum}
\usepackage{graphicx}
\usepackage{multirow}
\usepackage[noabbrev]{cleveref}
\usepackage[T1]{fontenc}
\usepackage{color}
\usepackage{bm}
\usepackage{float}

\ifCLASSOPTIONcompsoc
\renewcommand\thesubfigure{(\alph{subfigure})}

\usepackage[caption=false, font=normalsize, labelfont=sf, textfont=sf]{subfig}
\else
\usepackage[caption=false, font=footnotesize]{subfig}
\fi

\begin{document}
	
\title{Deep Discriminative Representation Learning with Attention Map for Scene Classification}

\author{
	Jun Li$^*$, Daoyu Lin$^*$, Yang Wang, Guangluan Xu, Chibiao Ding$^{\dag}$

	\thanks{Jun Li, Daoyu Lin, Yang Wang, Guangluan Xu, Chibiao Ding are with the Institute of Electronics, Chinese Academy of Sciences, Beijing, 100190, China (e-mail: lijun215@mails.ucas.ac.cn; lindaoyu15@mails.ucas.ac.cn; primular@163.com; gluanxu@mail.ie.ac.cn; cbding@mail.ie.ac.cn).
		
		Jun Li, Daoyu Lin, Yang Wang, Guangluan Xu are with the Key Laboratory of Network Information System Technology (NIST), Institute of Electronics, Chinese Academy of Sciences, Beijing, 100190, China
	
	Jun Li is with the School of Electronic, Electrical and Communication Engineering, University of Chinese Academy of Sciences, Beijing, 100190, China }
    \thanks{$^*$ Equal contribution  $^\dag$ Corresponding author}
}
\maketitle

\begin{abstract}
Learning powerful discriminative features for remote sensing image scene classification is a challenging computer vision problem.
In the past, most classification approaches were based on handcrafted features. However, most recent approaches to remote sensing scene classification are based on Convolutional Neural Networks (CNNs). The de facto practice when learning these CNN models is only to use original RGB patches as input with training performed on large amounts of labeled data (ImageNet). In this paper, we show class activation map (CAM) encoded CNN models, codenamed DDRL-AM, trained using original RGB patches and attention map based class information provide complementary information to the standard RGB deep models. To the best of our knowledge, we are the first to investigate attention information encoded CNNs. 
Additionally, to enhance the discriminability, we further employ a recently developed object function called "center loss," which has proved to be very useful in face recognition. Finally, our framework provides attention guidance to the model in an end-to-end fashion. Extensive experiments on two benchmark datasets show that our approach matches or exceeds the performance of other methods.

\end{abstract}


\IEEEpeerreviewmaketitle

\section{Introduction}
 
Remote sensing image scene classification, automatically extracting valuable information from each scene image and categorizing them into a discrete set of meaningful land use and land cover (LULC) classes, has become a research hotspot ~\cite{gomez2015multimodal,xia2017aid}. In recent years, a variety of methods have been put forward for image scene classification task for its wide application, such as geospatial object detection, urban planning, and land resource management. However, due to its unique spatial patterns, scene classification continues to be a challenging task~\cite{cheng2017remote}.

For traditional classification methods, the scene classification task generally consists of two steps: feature extraction and classification. A lot of research work focuses on how to describe image features better. At the earliest, most of the methods are based on per-pixel analysis due to the extremely low spatial resolution of remote sensing images. As the spatial resolution of the image increases, it is tough to classify scene images solely by pixel level~\cite{blaschke2001s}. Object-level approach~\cite{blaschke2008object} has led a long time for the task of remote sensing image analysis. Although the above two classification methods have superior performance, they can not show the semantic information of the image. Significant efforts have been made in developing sophisticated descriptors to capture the semantics of the scene.

\begin{figure}[t]
	\centering
	\subfloat[intra-class variations: palace]{
		\includegraphics[width= 0.15\textwidth]{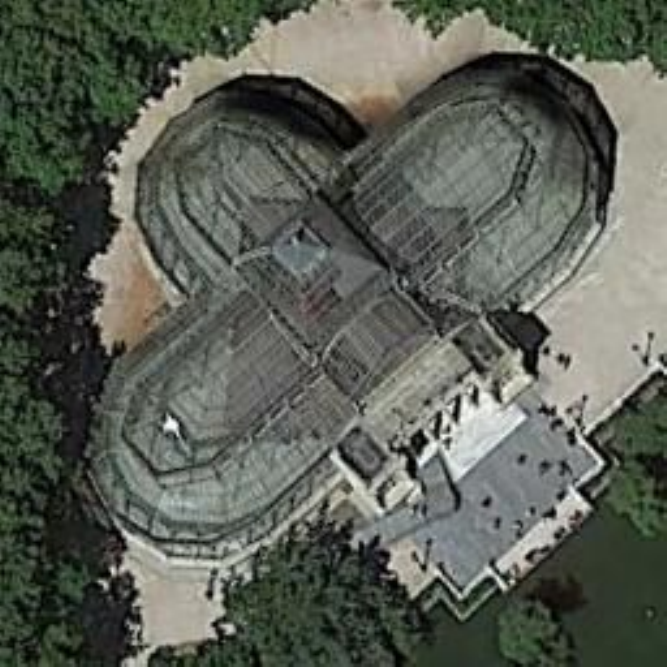}
		\includegraphics[width= 0.15\textwidth]{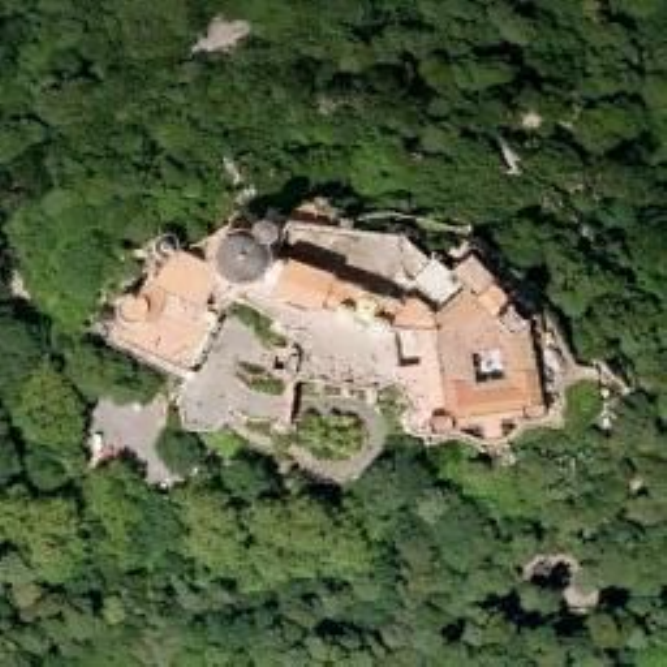}
		\includegraphics[width= 0.15\textwidth]{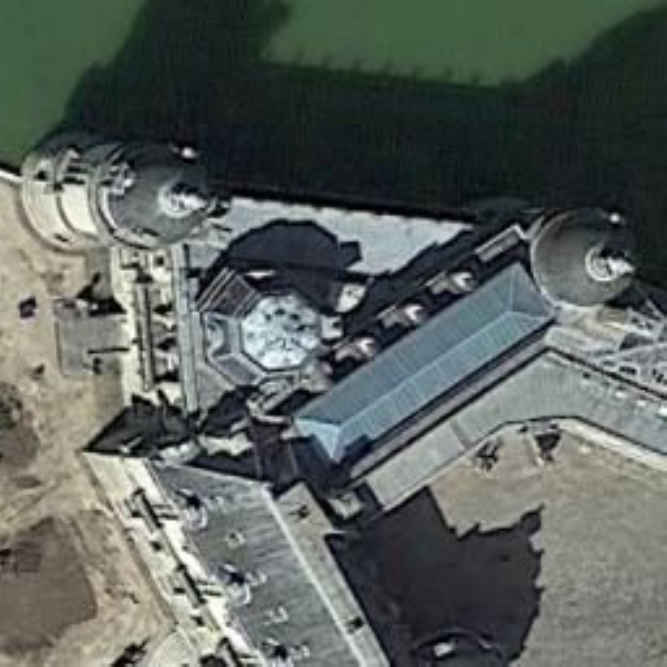}}
	\hfill
	\vspace{-0.5 em}
	\subfloat[small inter-class dissimilarity: freeway vs railway vs runway]{
		\includegraphics[width= 0.15\textwidth]{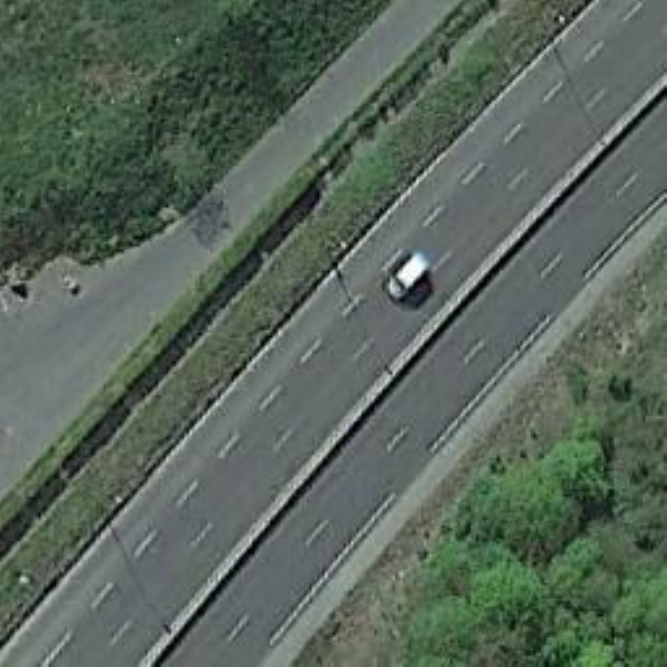}
		\includegraphics[width= 0.15\textwidth]{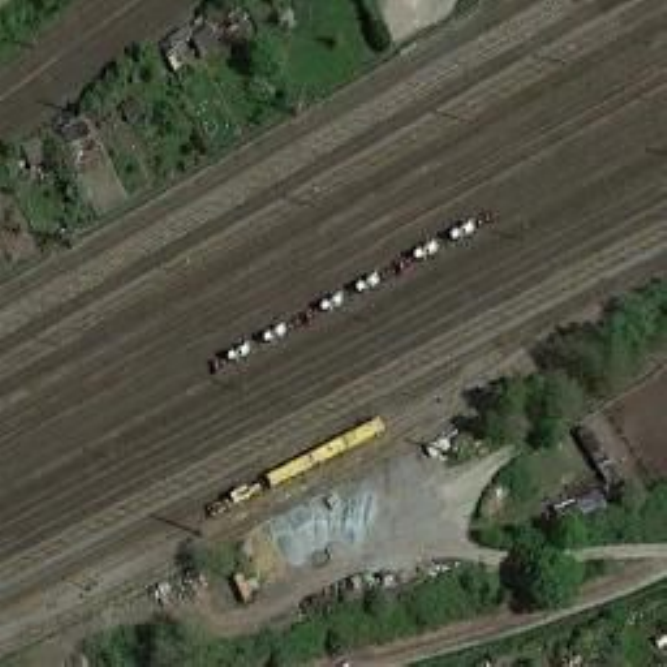}
		\includegraphics[width= 0.15\textwidth]{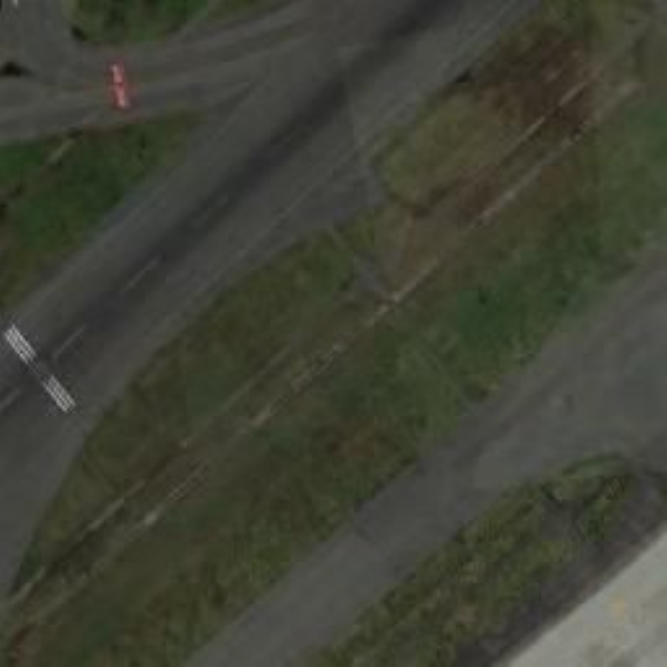}}
	\hfill
	\vspace{-0.5 em}
	\subfloat[Small object: airplane, baseball diamond, and golf course ]{
		\includegraphics[width= 0.15\textwidth]{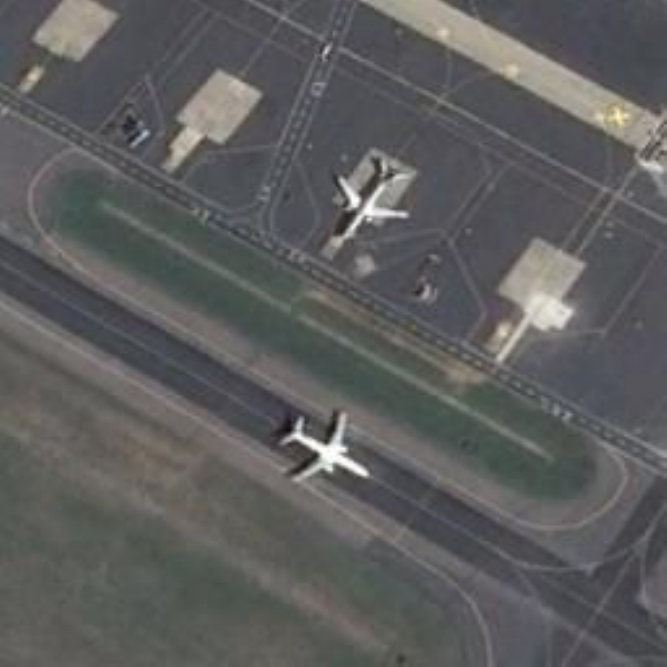}
		\includegraphics[width= 0.15\textwidth]{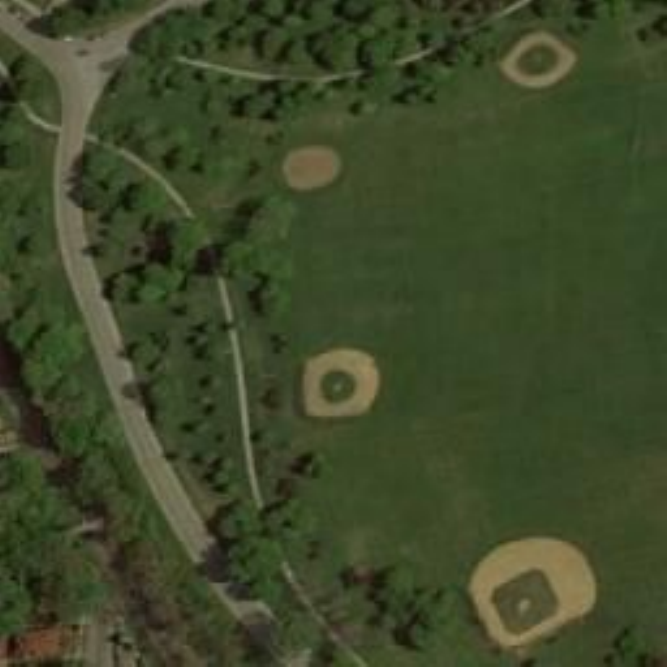}
		\includegraphics[width= 0.15\textwidth]{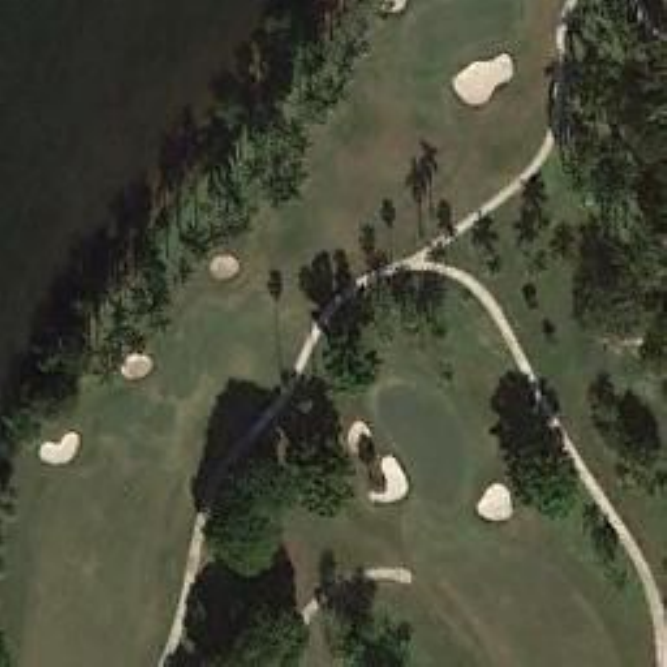}}
	\hfill
	\vspace{-0.5 em}
	\caption{Three major challenges: (a) large intra-class variations (b) small inter-class dissimilarity (c) small object in scene image.  These examples are from the challenging NWPU-RESISC45 dataset~\cite{cheng2017remote}}
	\label{problems} 
	\vspace{-0.5 em}
\end{figure}

With the deep Convolutional Networks (ConvNets) now being the architecture of choice for large-scale image recognition, a variety of CNN-based methods~\cite{hu2015transferring,zou2015deep} have been dominating the field of remote sensing image scene classification mainly due to its capacity to learn hierarchical representaion to describe the image scene.

Although deep learning-based approaches have been successfully developed in remote sensing field, three main factors make the task of remote sensing image scene classification more difficult, such as large intra-class variations, small inter-class dissimilarity and objects in the scene images are usually smaller and more separated than the natural image, as shown in Fig.~\ref{problems}. These problems lead to the performance of the scene classification cannot achieve the best. Learning robust and discriminative representation from remote sensing images is an urgent problem to be solved.

To overcome this limit, recently, distance metric learning, precisely measure both similarity and dissimilarity of features and labels between images, can learn a discriminative space. Wang \emph{et~al.}~\cite{wang2017learning} propose a discriminative distance metric learning method with label consistency~(LC). Firstly, they extracted the dense scale-invariant feature transformation features from remote sensing image. And then used spatial pyramid maximum pooling with sparse coding to encode the features. Cheng \emph{et~al.}~\cite{cheng2018deep} propose a simple but effective method to learn discriminative CNNs (D-CNNs) to boost the performance of remote sensing image scene classification. Different from the traditional CNN models that minimize only the cross-entropy loss, the D-CNN models are trained by optimizing a new discriminative objective function. The above two metric learning works are consistent with the purpose of our method, which is to map the same scene class closely to each other and different classes separable as farther apart as possible. But they do not distinguish between different areas of the input sample. Because we believe that the object area of the scene should be given a high weight, representing the unique characteristics of a particular category, which can effectively improve the classification performance.

Therefore, in order to solve the three challenges mentioned above, we propose a deep discriminative representation learning method with the attention map (DDRL-AM) to improve the performance of remote sensing image scene classification. 
Inspire of attention mechanism~\cite{vaswani2017attention}, we generate attention map (AM) for all images, where each pixel value indicates the importance of the corresponding original image. After that, the attention map is merged with the semantic features of the original image. Further, to better distinguish similar categories, apart from minimising the cross-entropy loss, we introduce the center loss~\cite{wen2016discriminative} to enforce the image from the same class as close as possible.

Our main contributions can be summarized as follows:

1) We propose a novel DDRL-AM learning scheme to address the problem of intra-class inconsistency and inter-class indistinction in the remote sensing image scene classification task, achieving the learned features to ensure intraclass compactness and interclass discriminability.

2) We generate attention map related to the original image from the pre-trained model as priors for the classification task, and make attention map an explicit input component of the end-to-end training for the first time, aiming to force the network to focus attention on the most discriminative parts.

3) We design a novel feature fusion pattern to effectively extract discernible features by combining the attention map and the original pixel space.

4) To enhance the discriminative power, we introduce a new loss function, which simultaneously learns a center for deep feature of each class and increases the decision boundary of different class.

The rest of this paper is organized as follows. Section II gives a brief review of related works.
Section III describes the proposed method in detail. Section IV introduces the experimental setup and presents results. The paper concludes in Section V with a summary and an outlook.

\section{RELATED WORK}
In this section, we will briefly review the development of remote sensing image scene classification, and specifically present the differences between existing methods and our proposed method.
\subsection{Feature Representation}
The early works mainly focus on handcrafted features and emerge a series of different types of feature descriptors, including global feature descriptors, (like color histograms~\cite{swain1991color}, texture features~\cite{haralick1973textural,jain1997object,ojala2002multiresolution}, GIST~\cite{oliva2001modeling}), and local feature descriptors, (like scale invariant feature transform (SIFT)~\cite{lowe2004distinctive}, histogram of oriented gradients(HOG)~\cite{dalal2005histograms}). These descriptors have proven successful in a wide variety of computer vision tasks. However, the limitation of these methods is that rely on low-level features and involve abundant engineering skills. And these methods cannot effectively capture the rich semantic information inside the remote sensing images.

To make up this deficiency of the above methods, unsupervised feature learning strategy, aiming to learn a set of basis functions used for feature encoding, is considered as an effective solution. One of the most popular methods is the bag of visual words (BoVW)~\cite{fei2005bayesian}, where the visual dictionaries (codebooks) are generated by conducting k-means clustering on the local features such as SIFT~\cite{lowe2004distinctive}. To learn more discriminative feature, histogram-based features were applied to the vector of locally aggregated descriptors (VLAD)~\cite{jegou2010aggregating} and Fisher Vectors~\cite{perronnin2010improving}. Unfortunately, there are two major shortcomings in these methods. On the one hand, the above method ignores the spatial distribution information of visual words. Therefore, a large number of spatial pyramid-based methods have been proposed, for example, the classic method is the spatial pyramid match kernel (SPMK)~\cite{lazebnik2006beyond} and the spatial co-occurrence kernel (SCK)~\cite{yang2010bag}. On the other hand, the lack of data class information cannot guarantee the better discernible ability of learned feature. Accordingly, supervised deep learning methods significantly improve performance.

The remarkable performance of AlexNet~\cite{krizhevsky2012imagenet}, which was the winner of ImageNet large scale visual recognition challenge (ILSVRC)~\cite{russakovsky2015imagenet} in 2012, promotes a large amount of visual research work to deep learning. And convolutional neural networks (CNNs) have been successfully applied to the scene classification. To solve the scarcity of remote sensing training data, the existing deep learning-based has three kinds of training modalities: the usual training from scratch~\cite{penatti2015deep}, directly feature extraction from pre-trained model~\cite{cheng2017remote} and fine-tuning of pre-trained networks~\cite{castelluccio2015land,zhang2016deep}. Similarly, our approach is also based on existing convolution models, and also involves fine-tuning strategies. Deep learning features are not only automatically learned from raw data via deep-architecture neural networks, but also can learn more powerful feature representations of data with multiple levels of abstraction.

All the methods mentioned above, whether using hand-crafted features or automatically learning feature via unsupervised and supervised method, not aware that only discriminative regions are essential for identifying different scene tasks in the entire image area.

\subsection{Attention Mechanism}
More recently, some research works have attempted to focus the feature extraction process on discriminative image regions. 
Saliency map is currently widely used for spatial semantic segmentation~\cite{zhou2018weakly,wei2017object} and spatial object localization~\cite{zhang2018top}. However, saliency has been less exploited for improving model classification. In the context of image classification, Jetley \emph{et~al.}~\cite{jetley2018learn} propose an end-to-end trainable attention module for convolutional neural network architectures, the primary purpose is to amplify the relevant and suppress the irrelevant or misleading. Furthermore, in~\cite{wang2017residual}, residual attention network is proposed, where stack attention modules generate attention-aware features. Their system is robust against noisy labels. In ~\cite{woo2018cbam,park2018bam}, researchers propose a simple and effective attention module, which infers an attention map along two separate pathways, channel and spatial. The attention maps are multiplied to the input feature map for adaptive feature refinement. In order to focus on the channel relationship, Hu \emph{et~al.}~\cite{hu2017squeeze} propose a novel architectural unit, namely, "squeeze-and excitation" block, which adaptively recalibrates channel-wise feature responses. Nakka \emph{et~al.}~\cite{nakka2018deep} incorporate attention maps and structured representation into a deep learning framework. Specifically, the attention weight is introduced to the VLAD~\cite{jegou2010aggregating}  vector. However, while structured attentional representation outperforms others, this process is tedious. In contrast to existing approaches, our approach also makes use of the attention map. However, it is fundamentally different from above object recognition approaches because we merely use the attention map as input to the new branch. 

\subsection{Feature Aggregation}
Feature fusion~\cite{simonyan2014two} is an important method in pattern recognition, which can comprehensively utilize multiple image features to achieve complementary advantages of multiple features and obtain more robust and accurate recognition results. And feature fusion ideas are often used in deep learning network models. Simonyan \emph{et~al.}~\cite{simonyan2014two} first proposed a deep convolutional neural network model using a dual-flow architecture to solve the motion recognition problem in video. The model establishes a spatial stream convolutional neural network and a time-flow convolutional neural network for independent training. The final Softmax classification output layer fuses the two networks and belongs to the classifier level fusion. On the basis of this, Feichtenhofer \emph{et~al.}~\cite{feichtenhofer2016convolutional} improve the network fusion method, and proposed a spatial feature fusion method and a temporal feature fusion algorithm, which can be fused not only in the Softmax layer but also in the ReLU layer after the convolution layer to achieve feature level fusion. In particular, Chaib \emph{et~al.}~\cite{chaib2017deep} propose a method based on discriminant correlation analysis to feature fusion. Our strategy is to use different branch structures to extract the features of the original image and the attention map separately for feature fusion.

\section{Method}

\begin{figure*}[t]
	\centering
	\includegraphics[width=\linewidth]{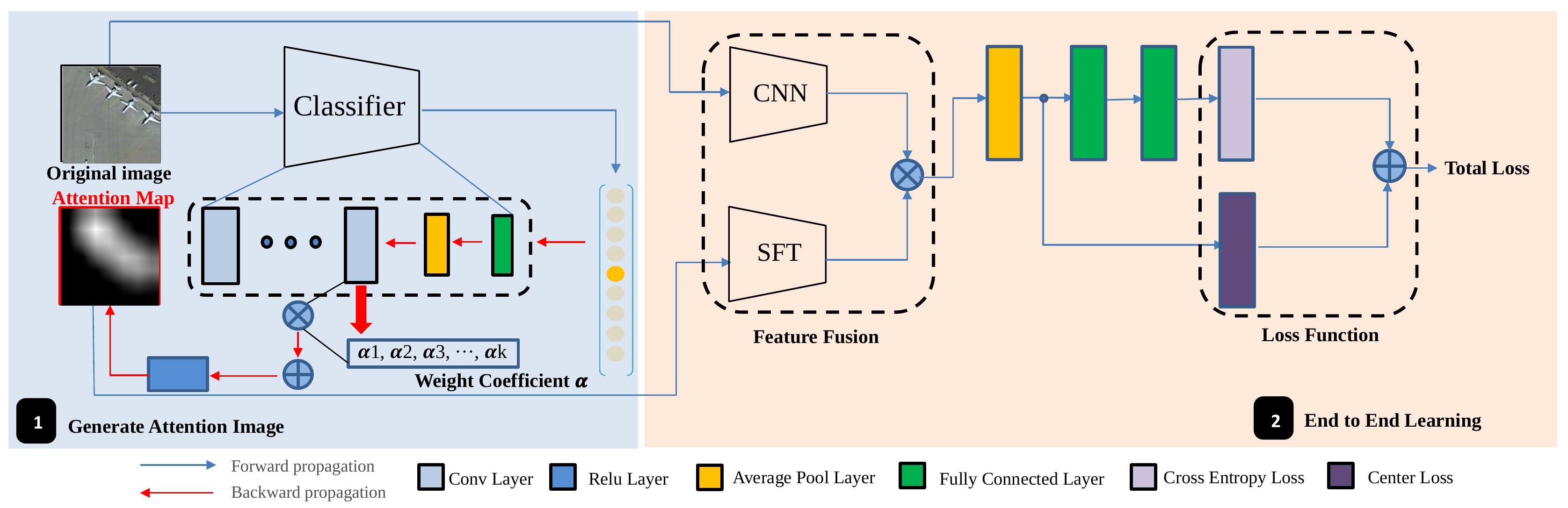}
	\caption{Overview of our framework, which consists of 1) pre-trained CNN network from Imagenet dataset and Grad-CAM architecture to generate discriminative attention map, 2) an inconsistent two-stream architecture joint optimization to fuse high-feature from original image and attention map and combine multiple loss functions. For simplicity, the different color block represents different network structure layer respectively.}
	\label{model} 
\end{figure*}

As shown in Fig.~\ref{model}, the proposed Deep Discriminative Representation Learning with Attention Map (DDRL-AM) consists of two novel components: the attentional  generative network and the deep feature fusion model. In this section, we first introduce our novel attentional representation learning framework and then elaborate each of them in details.

\begin{algorithm}[t]
	\caption{DDRL-AM }
	\textbf{Step1}  Generate Attention Map \ \\
	\KwIn{Full Image $X$ \ \\}
	\KwOut{Full Attention Map $X_{am}$\ \\}
	Fine-tune ResNet-18 model on training data sets\ \\
	Forward Inference Full Image $X$\ \\
	Calculate Weight Coefficient $\alpha$\ from Eqn.~\ref{eq:gradient}\ \\
	Obtain Gray Scale Saliency Map $X_{sm}$ from Eqn.~\ref{eq:location} \ \\
	Return Attention Map $X_{am}$ by upsampling $X_{sm}$ to the $X$ size  \ \\
	\textbf{Step2}  End-to-End Learning\ \\
	\KwIn{Full image $X$, Full Attention Map $X_{am}$\ \\}
	\KwOut{Predict Probability $P$\ \\}
	\While{Epoch =1,2,\ldots,Num}
	{
		Take $X$,$X_{am}$ \ \\
		$P$ = $f$($X$,$X_{am}$) \ \\
		Calculate $L_{total}$ from Eqn.~\ref{eq:ltotal} \ \\
		$BP$($L_{total}$) get gradient $w.r.t$ $\bm{\theta}$ \ \\
		Update $\bm{\theta}$ using ADAM
	}
\end{algorithm}

\subsection{Overall Architecture}
\label{overar}

The left part of our framework (Fig.~\ref{model}) shows the structure we generate the attention map, we train the off-the-shelf ResNet-18~\cite{he2016deep} from fine-tuning, because the features learned are more oriented to remote sensing images after fine-tuning. Then we generate attention map for all images by Gradient-weighted Class Activation Mapping (Grad-CAM) algorithm (the section~\ref{sec:attentionmap}). The right half of our framework (Fig.~\ref{model}) shows our trainable  CNN model structure, including feature fusion of different branch structures (the section~\ref{sec:featurefusion}) and integration of two loss function (the section~\ref{sec:lossfunction}). The overall algorithm is summarized in Algorithm $1.$


\subsection{Attention Map}
\label{sec:attentionmap}
Most of the Attention Map (AM) in the current literature~\cite{Zhou_2016_CVPR,selvaraju2017grad,chattopadhay2018grad} is mainly generated to explain the prediction corresponding network model. And the saliency image summarizes where a DNN looks into an image for recognizing their predictions.

\begin{figure}[h]

	\centering
	\includegraphics[width=0.23\textwidth]{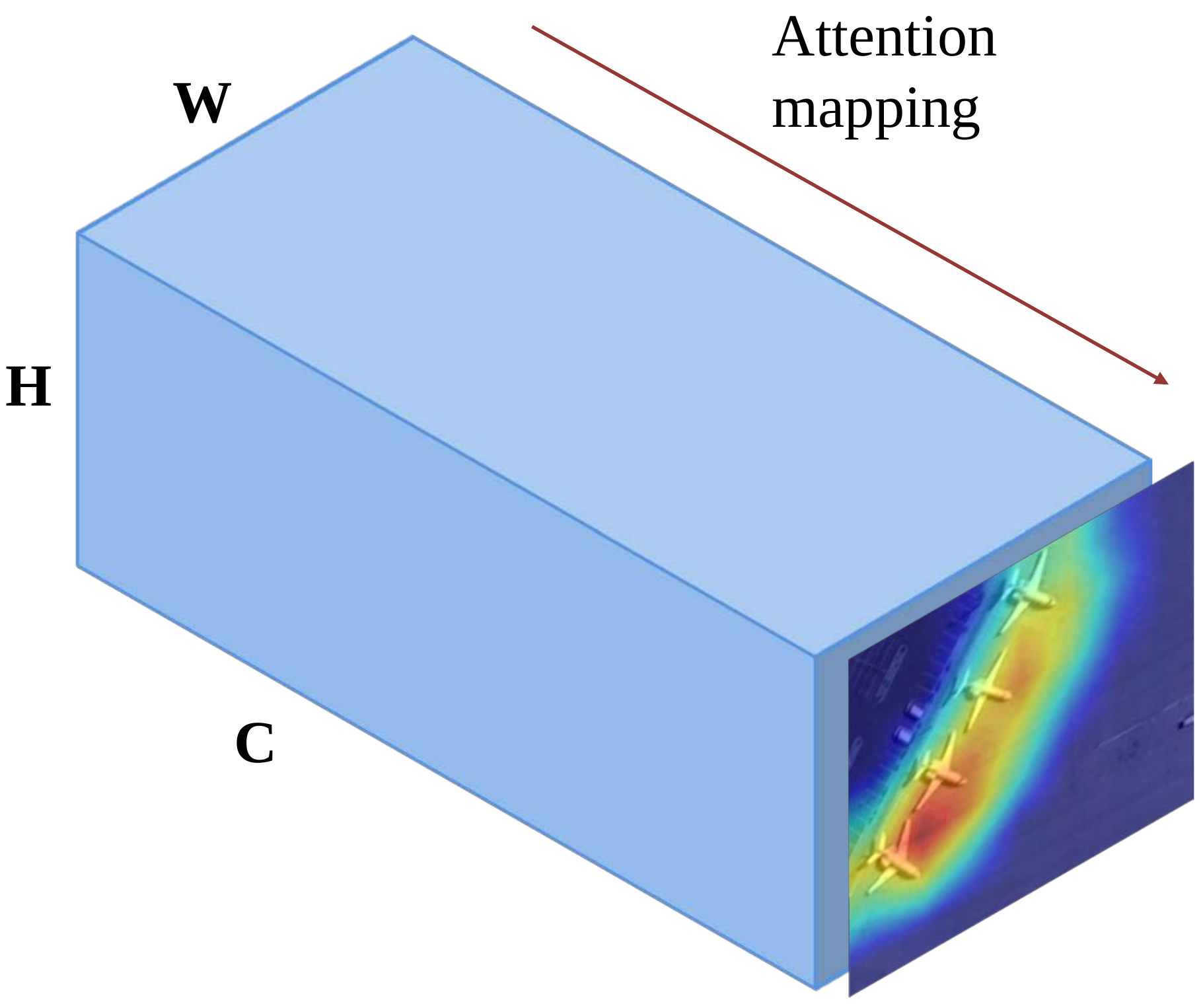}
	\vspace{-0.5em}
	\caption{ Attention mapping over conv\_5 feature maps from fine-tuned resnet-18 }
	\label{weighfeaturemap}
\end{figure}

Recently, three commonly used techniques to compute attention maps given category labels CAM~\cite{Zhou_2016_CVPR}, Grad-CAM~\cite{selvaraju2017grad}, and Grad-CAM++~\cite{chattopadhay2018grad}.
CAM shows that the average pooling layer can help to generate attention maps w.r.t specific class. Grad-CAM extends CAM to many different available architectures for tasks like image captioning and VQA, which help to explain model decisions.
We do not use CAM because it is inflexible, requiring architecture modification for generating trainable attention. While Grad-CAM++ proposes a better class activation map by modifying the way weights are computed, its high computational cost in calculating the second and third derivatives makes it impractical to generate saliency map.

Lots of milestone work~\cite{zeiler2014visualizing} assert that deeper representations in a CNN capture higher-level visual concept. Furthermore, the neurons in last convolutional layers look for semantic class-specific information in the image.

In this paper, we resort to Gradient-weighted Class Activation Mapping (Grad-CAM) method~\cite{selvaraju2017grad} to produce attention map image on all training and test datasets. 


In order to obtain the grey-scale image, which one-to-one correspondence with the original image, the framework we proposed contains two-step. In other words, forward propagation and backward propagation. As shown in Fig.~\ref{model}, for forward propagation, we use ResNet~\cite{he2016deep} as a base recognition model, by fine-tuning the pre-trained networks on remote sensing training data. For backward propagation, we mainly use the Grad-CAM to generate the grey-scale image. The specific implementation process is as follows. We firstcompute the neuron importance weights $\alpha_k^c$:
\begin{equation}
\alpha_k^c = \frac { 1 } { Z } \sum _ { i } \sum _ { j }  \frac { \partial y ^ { c } } { \partial A _ { i j } ^ { k } }
\label{eq:gradient}
\end{equation}
In Eq.~\ref{eq:gradient}, $y ^ { c }$ (before the softmax) denotes the score for class $c$, $A  ^ { k }$ denotes the feature maps in last convolutional layer. The $\alpha_k^c$ represents the relative importance coefficient of feature map $k$ for a target class $c$. 

Then, we obtain the class-discriminative localization map $L _ { \mathrm { Grad } -\mathrm { CAM } } ^ {c}$ (Eq.~\ref{eq:location}). Note that the result is the same size as the convolutional feature maps(7*7), we need to further upsample the operation to restore the saliency map that is the same size as the original image.
\begin{equation}
L _ { \mathrm { Grad } -\mathrm { CAM } } ^ {c} = \operatorname {ReLU}\left(\sum _ { k } \alpha _ { k } ^ { c } A ^ { k } \right)
\label{eq:location}
\end{equation}

Essentially, the Grad-CAM algorithm constructs a spatial attention map by weighing the different feature maps of the last convolutional layer of the fine-tuned model(Fig.~\ref{weighfeaturemap}).

\begin{figure*}[t]
	\centering
	\includegraphics[width=0.8\textwidth]{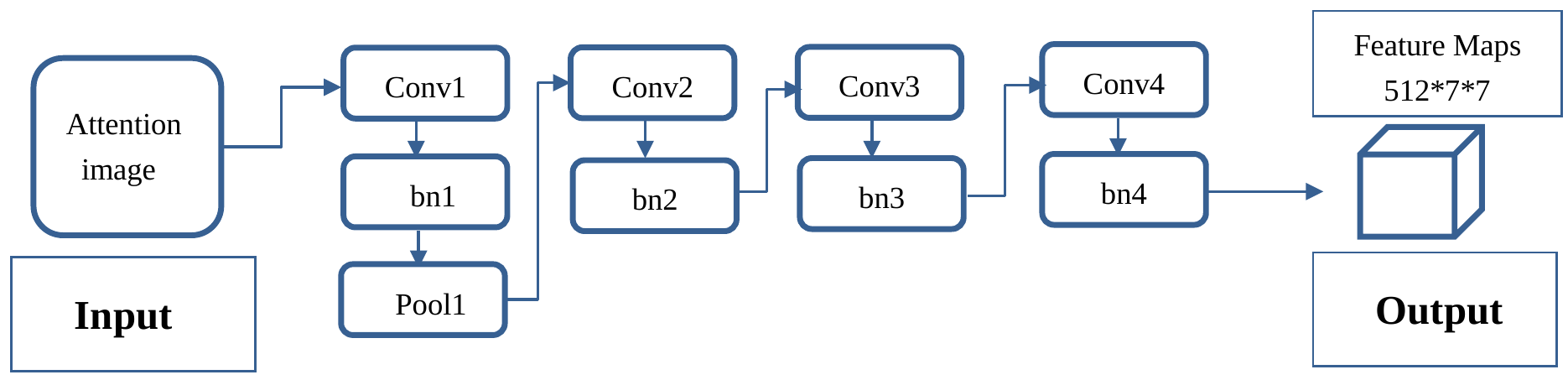}
	\caption{spatial feature transformer architecture}
	\vspace{2mm}
	\label{sft}
\end{figure*}
\vspace{1mm}

\begin{table*}[!ht]
	\centering
	\scriptsize
	\caption{SFT Architecture Parameter Setting}
	
	\begin{tabular}{|l|c|c|c|c|c|c|c|c|c|}
		
		\hline
		
		\multirow{2}{*}{Architecture} & \multicolumn{3}{c|}{Group1}                                                                                                                                                                          & \multicolumn{2}{c|}{Group2}                                                                                                       & \multicolumn{2}{c|}{Group3}                                                                                                            & \multicolumn{2}{c|}{Group4}                                                                                       \\ \cline{2-10} 
		
		& \begin{tabular}[c]{@{}c@{}}Conv1\\ (k\_num/k\_size/s/p)\end{tabular} & \begin{tabular}[c]{@{}c@{}}BN1\\ (k\_num/mom)\end{tabular} & \begin{tabular}[c]{@{}c@{}}Maxpool1\\ (k\_size/s/p)\end{tabular} & \begin{tabular}[c]{@{}c@{}}Conv2\\ (k\_num/k\_size/s/p)\end{tabular} & \begin{tabular}[c]{@{}c@{}}BN2\\ (k\_num/mom)\end{tabular} & \begin{tabular}[c]{@{}c@{}}Conv3\\ (k\_num)\end{tabular} & \multicolumn{1}{l|}{\begin{tabular}[c]{@{}l@{}}BN3\\ (k\_num)\end{tabular}} & \begin{tabular}[c]{@{}c@{}}Conv4\\ (k\_num)\end{tabular} & \begin{tabular}[c]{@{}c@{}}BN4\\ (k\_num)\end{tabular} \\ \hline
		
		\multirow{2}{*}{Parameter}    & \multirow{2}{*}{64/7*7/2*2/3*3}                                      & \multirow{2}{*}{64/0.1}                                    & \multirow{2}{*}{3*3/2/1}                                         & \multirow{2}{*}{128/3*3/2*2/1*1}                                     & \multirow{2}{*}{128/0.1}                                   & \multirow{2}{*}{256}                                     & \multirow{2}{*}{256}                                                        & \multirow{2}{*}{512}                                     & \multirow{2}{*}{512}                                   \\
		
		&                                                                      &                                                            &                                                                  &                                                                      &                                                            &                                                          &                                                                             &                                                          &                                                        \\ \hline
		
	\end{tabular}
	\label{sft para}
\end{table*}

\subsection{Feature Fusion}
\label{sec:featurefusion}
The feature fusion is an efficient step in scene understanding. In this section, we propose a simple and effective two-stream deep fusion architecture for remote scene classification. As shown in Fig.~\ref{model}, the first stream is called the original RGB stream, which feeds original RGB images into the network. The second stream is called saliency stream, using the grey images as input to the network. And we design spatial feature transformer (SFT) netwrok to extract valuable information from grey images. 

Note that the two branches respectively adopt different network structure to extract features. The RGB stream is consistent with the ResNet-18 network structure. The architecture of SFT is presented in Fig.~\ref{sft} and the specific parameter settings are shown in Table.~\ref{sft para}. 

SFT contains 4 convolutional layers and 4 batch normalization layers, and one max-pooling layer is only used following the first convolutional layer. The first convolutional layer has 64 filters of size 7x7 with a stride of 2 pixels and padding with 3 pixels. The stride and padding of other convolutional layers are set as 2 and 1 pixel respectively. The second, third, and fourth convolutional layers have 128, 256 and 512 filters with the size of 3x3 respectively. The batch normalization layers are consistent with the convolution kernel of the convolutional layer they are connected to. Max-pooling is carried out over a 3x3 window with stride 2.

We use a spatial feature fusion algorithm to fuse the two stream feature maps of the convolutional layer output. In particular, we use multiplicative fusion functions (Eq.~\ref{eq:mul}) for high-dimensional feature fusion.

\begin{equation}
y_d^{mul} = X_{i,j,d}^{rgb} * X_{i,j,d}^{grey}
\label{eq:mul}
\end{equation}
In Eq.~\ref{eq:mul}, $i,j \in [7,7]$, and $d$ equals 512. The number of channels in $y$ is still 512.

\subsection{Loss Function}
\label{sec:lossfunction}
Generally speaking, the cross-entropy is a loss function used frequently in classification problems, the softmax loss function is presented as follows.
\begin{equation}
L_{s}=-\sum\limits_{i=1}^m log \dfrac {e^{W_{y_i}^Tx_i+b_{y_i}}}{\sum_{j=1}^ne^{W_j^Tx_i}+b_j}.
\label{eq:ls}
\end{equation}
In Eq.~\ref{eq:ls}, $x_i \in {\mathbb R}^d$ denotes the ith deep feature, belonging to the $y_i$th class. $d$ is the feature dimension. $W_j \in {\mathbb R}^d $ denotes the $j$th column of the weights $W \in {\mathbb R}^{d \times n}$in the last fully connected layer and $b \in {\mathbb R}^n$ is the bias term. $m$ is size of mini-batch and $n$ is the number of class.  

Although softmax loss can directly address the classification problems, the deeply learned features trend to be separable but not be discriminative. Consequently, it is not suitable to directly use these features for recognition. To enhance the discriminative power of the deeply learned features in neural networks, Wen \emph{et~al.}~\cite{wen2016discriminative} first attempt to train the learning of CNNs by the joint supervision of softmax loss and center loss for face recognition task. Remote sensing scene classification tasks are similar to face recognition tasks, which needs identifiable features to distinguish between different categories.

A central goal is to minimize the intra-class variations while keeping the features of different classes separable, center loss function is proposed, as formulated in Eq.~\ref{eq:lc}.
\begin{equation}
L_{c}=\dfrac{1}{2}\sum\limits_{i=1}^m\lVert x_i-c_{y_i}\rVert_2^2.
\label{eq:lc}
\end{equation}
The $c_{y_i} \in {\mathbb R}^d$denotes the $y_i$th class center of deep features. The formulation effectively characterizes the intra-class variations.

However, the stochastic gradient descent (SGD)~\cite{bottou2012stochastic} optimizes the CNNs based on mini-batch, which not take the entire training set into account. To solve this problem, some modifications to be need. In each iteration, the centers are computed by averaging the features of the mini-batch class. However, not all categories are predicted correctly, to avoid large perturbations caused by few mislabelled samples, a scalar $\alpha$ is introduced to control the learning rate of the centers.
We employ the joint supervision of softmax loss and center loss to train the CNNs for discriminative feature learning. The formulation is given in Eq.~\ref{eq:ltotal}.
\begin{equation}
\begin{aligned}
L_{total}&=L_{s}+\lambda L_{c}\\&= -\sum\limits_{i=1}^m log \dfrac {e^{W_{y_i}^Tx_i+b_{y_i}}}{\sum_{j=1}^ne^{W_j^Tx_i}+b_j}+\dfrac{\lambda}{2}\sum\limits_{i=1}^m\lVert x_i-c_{y_i}\rVert_2^2.
\end{aligned}
\label{eq:ltotal}
\end{equation}
Where $\lambda$ is hyper-parameters. Finally, With the joint supervision, not only the inter-class features differences are enlarged, but also the intra-class features variations are reduced.
\begin{figure*}[t]
	\centering
	\subfloat[]{
		\includegraphics[width= 0.135\textwidth]{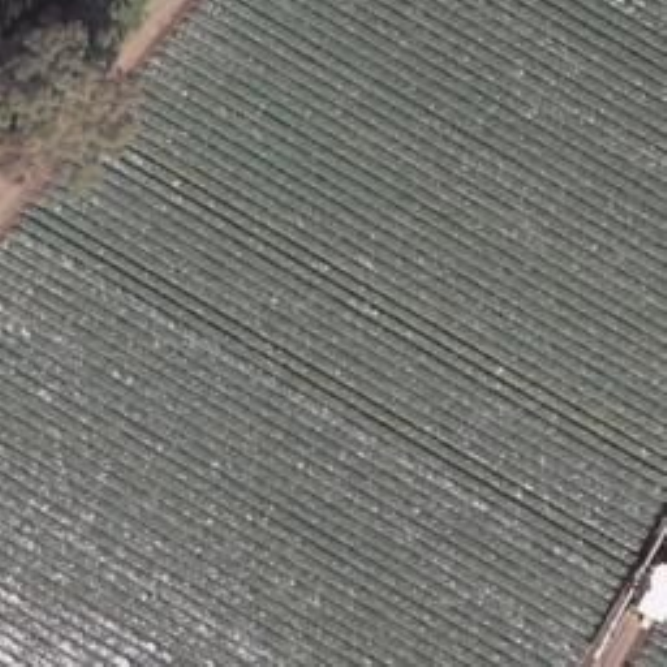}}
	\subfloat[]{
		\includegraphics[width= 0.135\textwidth]{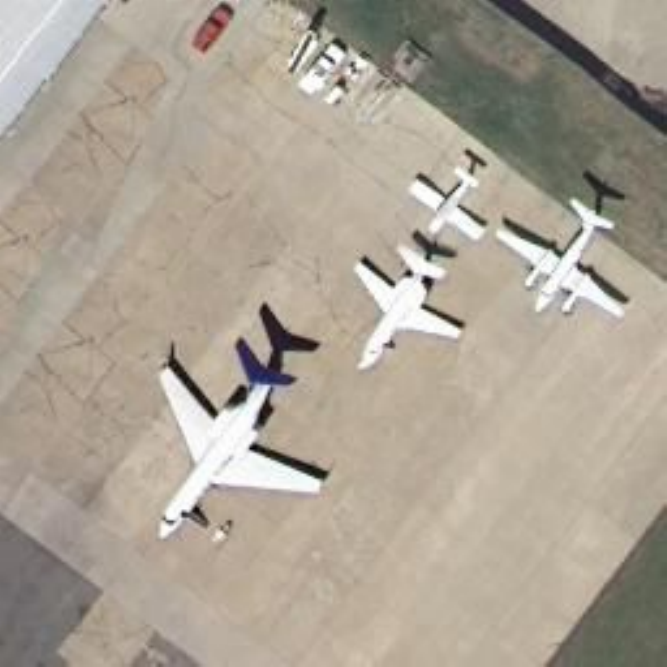}}
	\subfloat[]{
		\includegraphics[width= 0.135\textwidth]{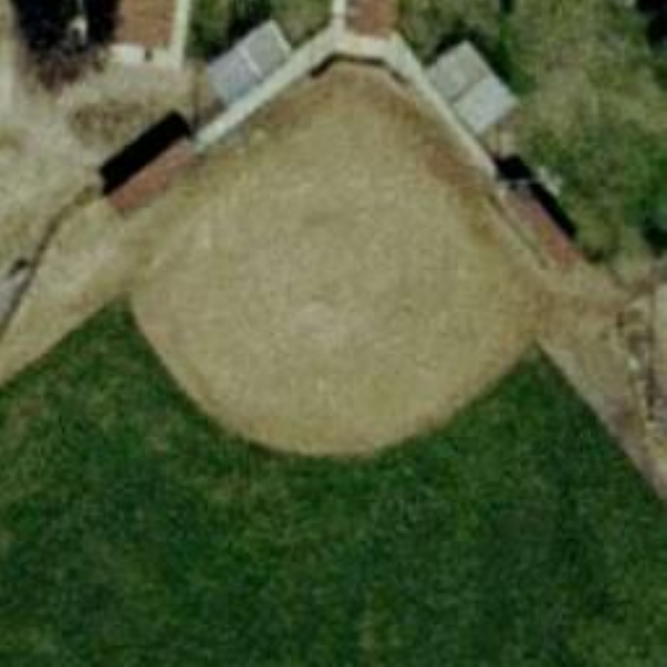}}
	\subfloat[]{
		\includegraphics[width= 0.135\textwidth]{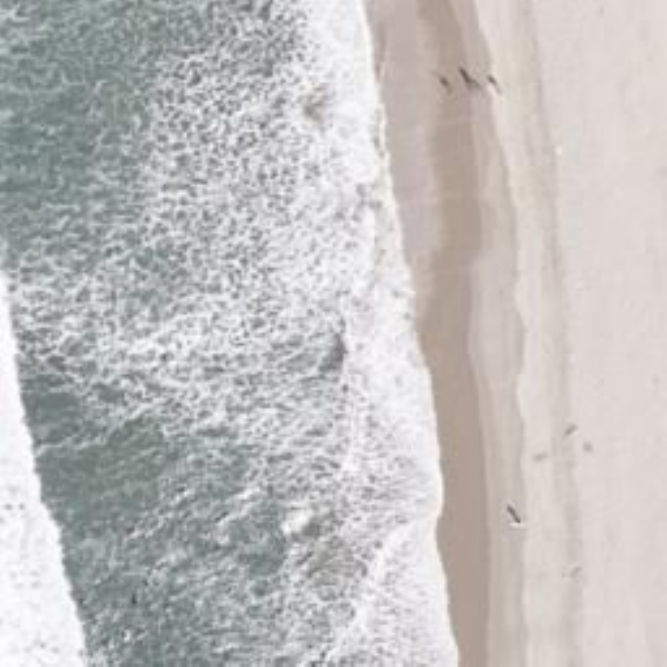}}
	\subfloat[]{
		\includegraphics[width= 0.135\textwidth]{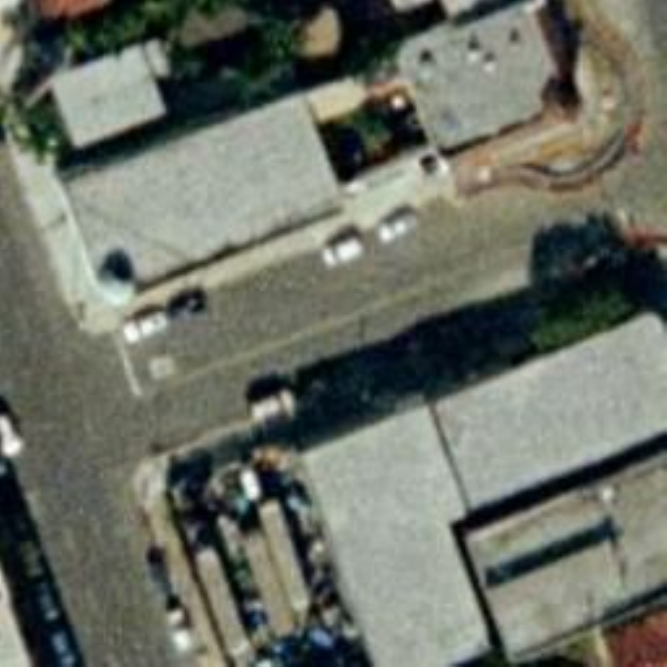}}
	\subfloat[]{
		\includegraphics[width= 0.135\textwidth]{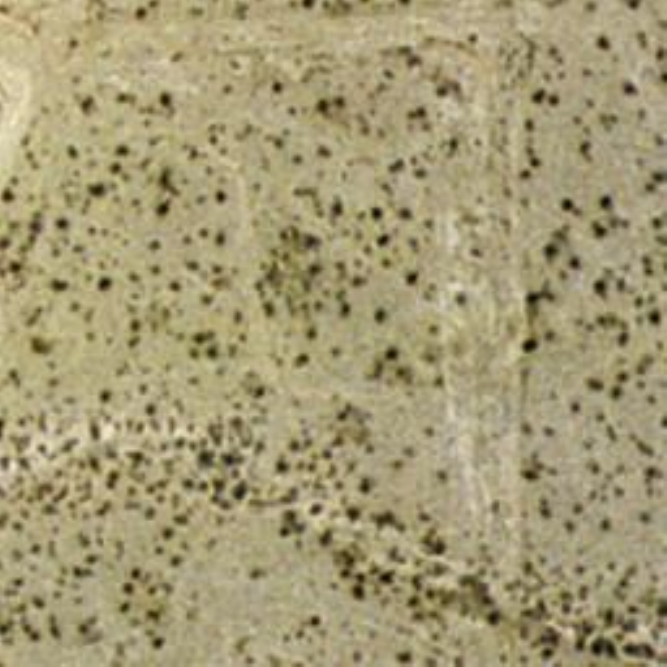}}
	\subfloat[]{
		\includegraphics[width= 0.135\textwidth]{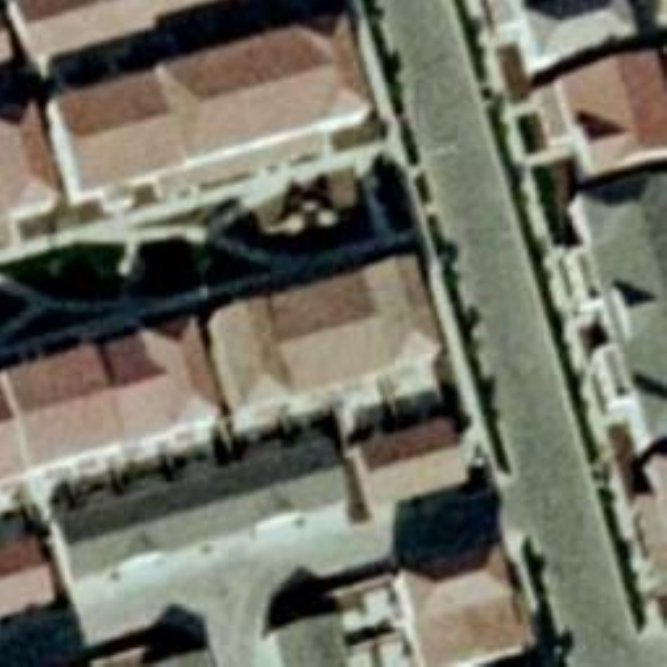}}
	\vspace{-0.5em}
	\hfill
	\subfloat[]{
		\includegraphics[width= 0.135\textwidth]{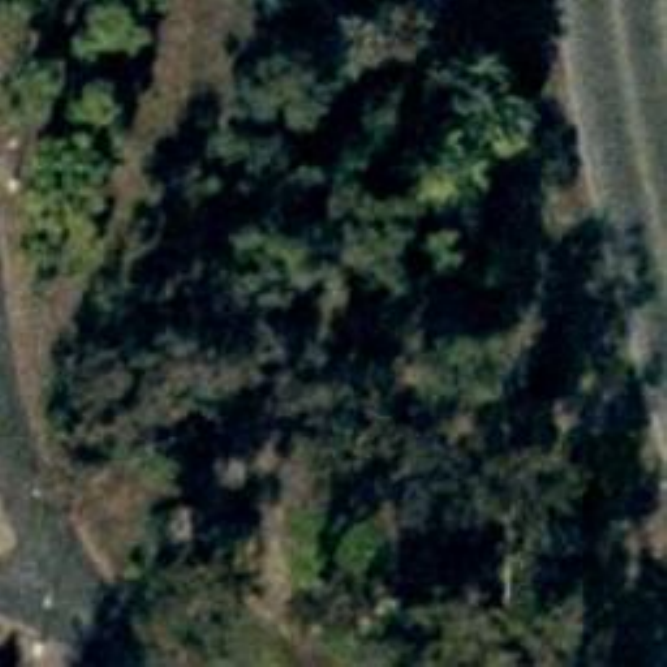}}
	\subfloat[]{
		\includegraphics[width= 0.135\textwidth]{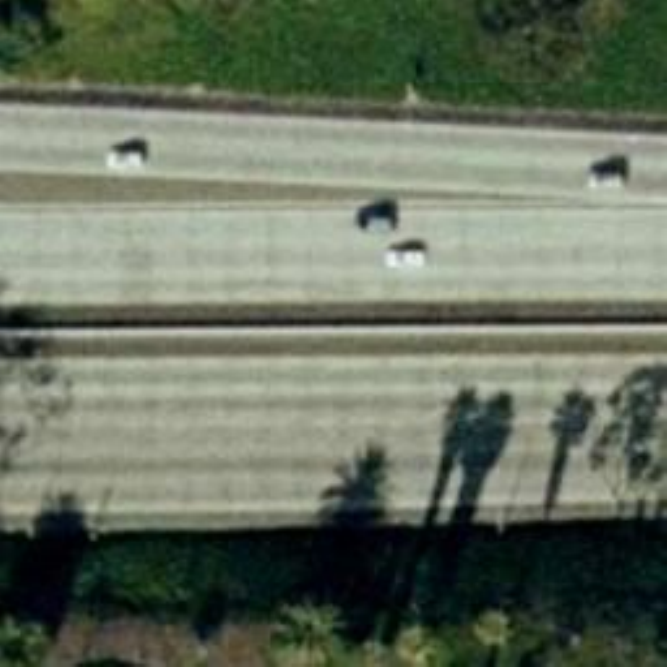}}
	\subfloat[]{
		\includegraphics[width= 0.135\textwidth]{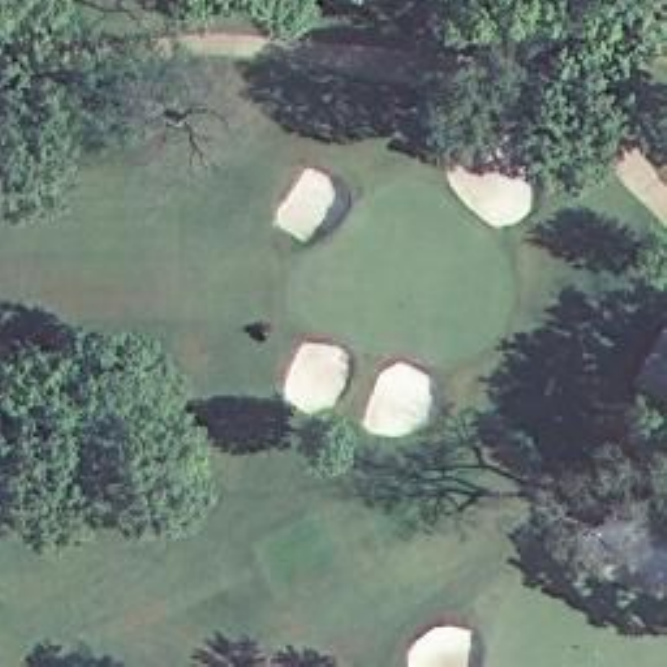}}
	\subfloat[]{
		\includegraphics[width= 0.135\textwidth]{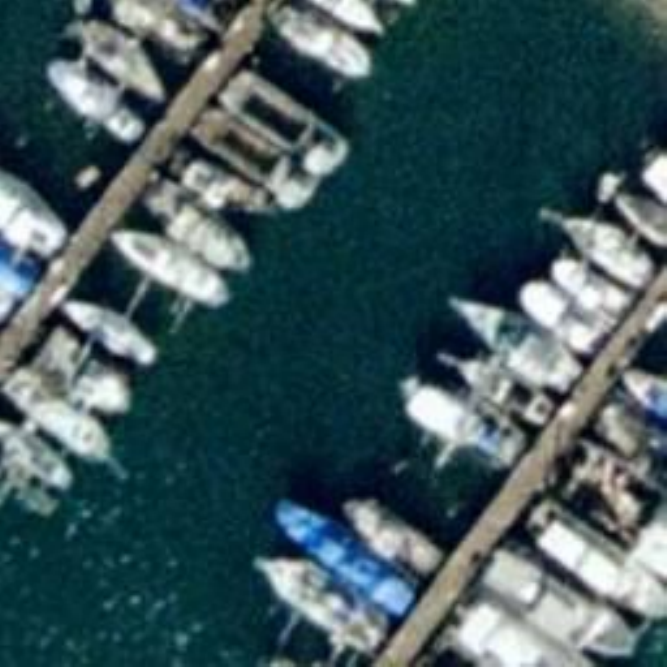}}
	\subfloat[]{
		\includegraphics[width= 0.135\textwidth]{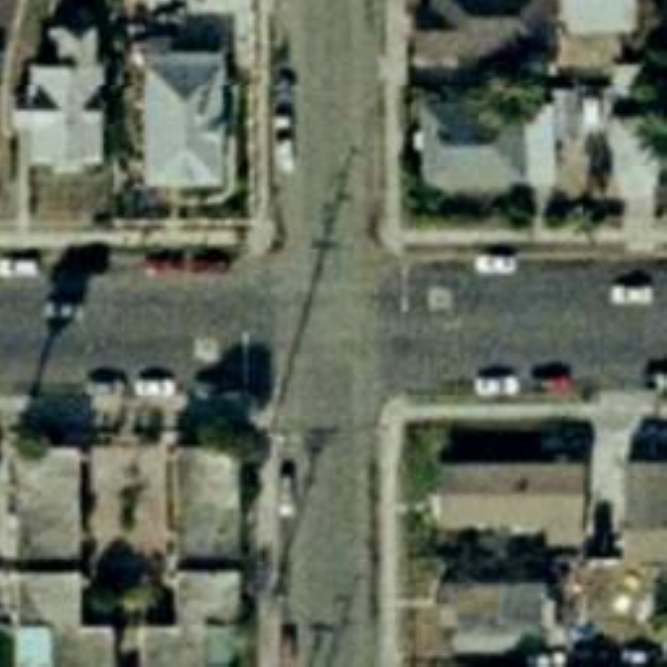}}
	\subfloat[]{
		\includegraphics[width= 0.135\textwidth]{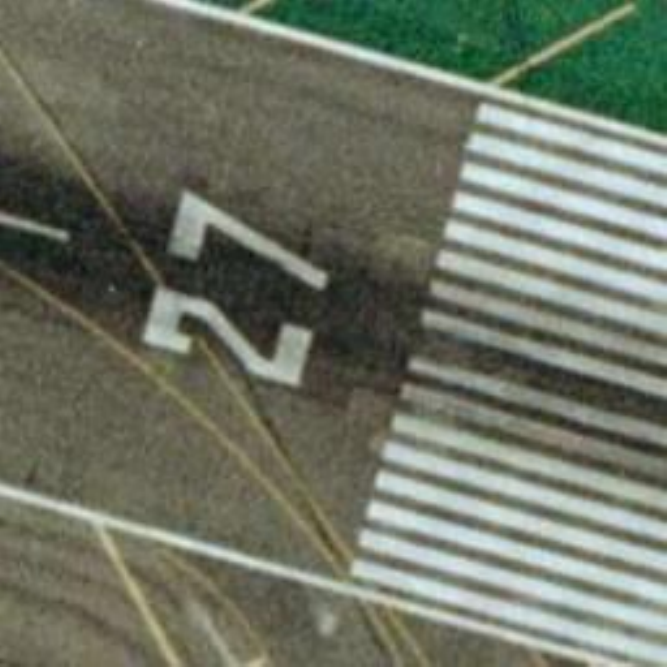}}
	\subfloat[]{
		\includegraphics[width= 0.135\textwidth]{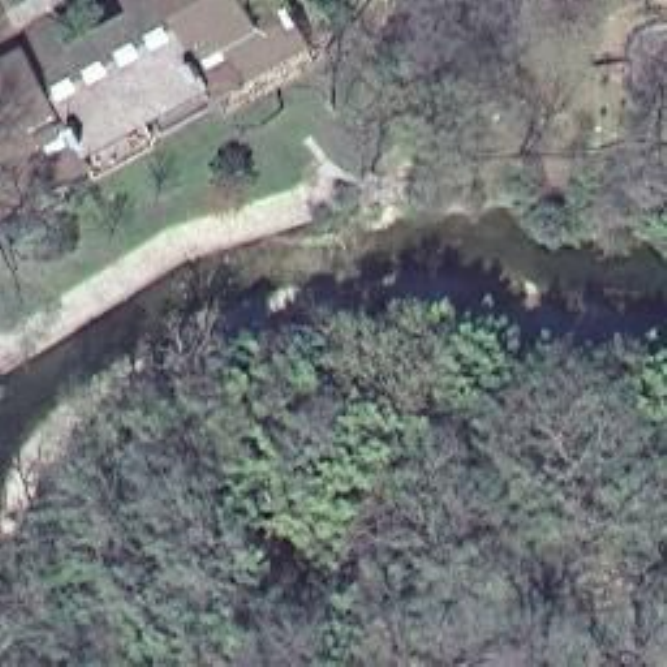}}
	\vspace{-0.5em}
	\hfill
	\subfloat[]{
		\includegraphics[width= 0.135\textwidth]{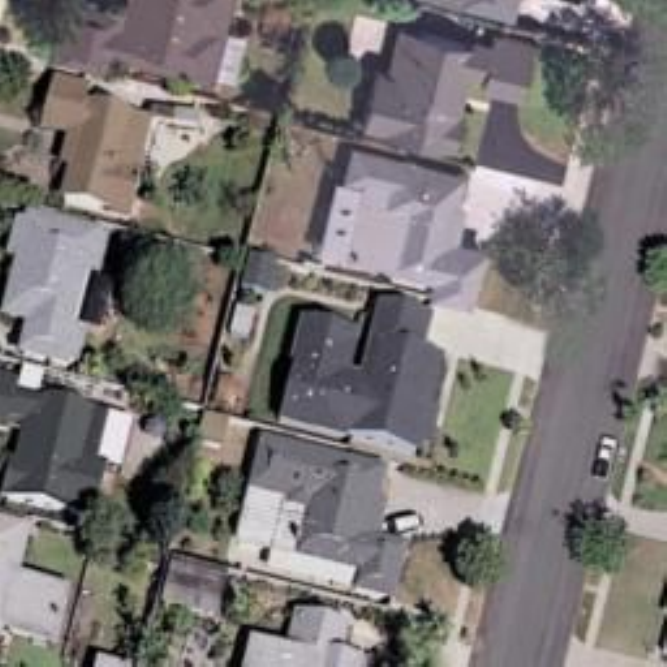}}
	\subfloat[]{
		\includegraphics[width= 0.135\textwidth]{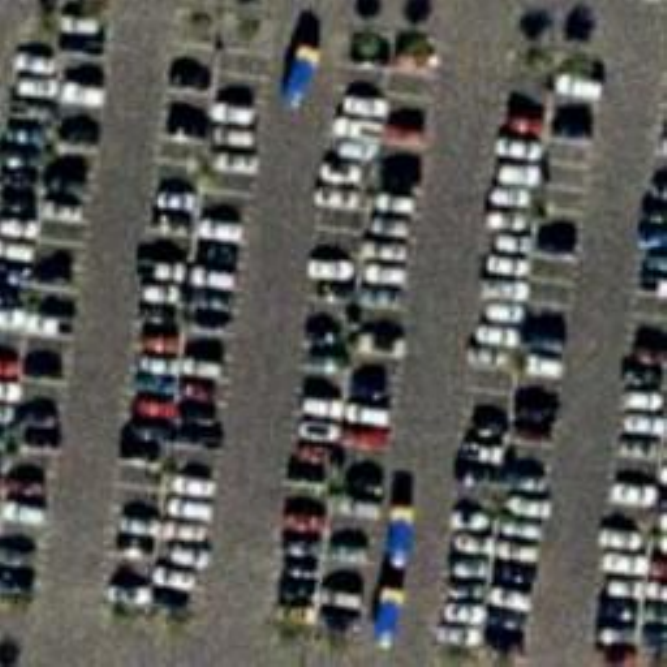}}
	\subfloat[]{
		\includegraphics[width= 0.135\textwidth]{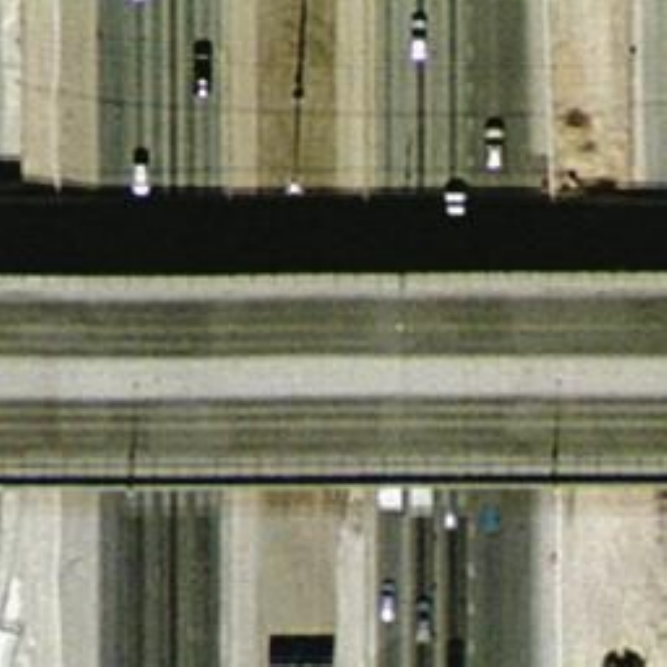}}
	\subfloat[]{
		\includegraphics[width= 0.135\textwidth]{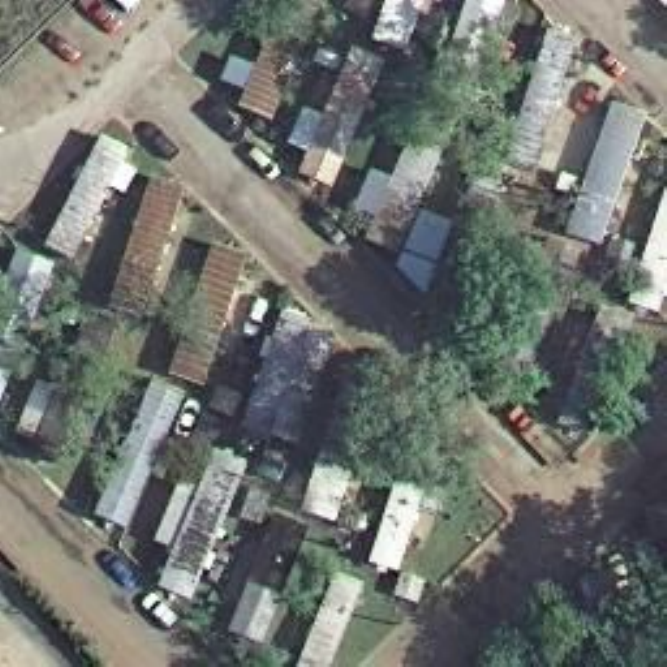}}
	\subfloat[]{
		\includegraphics[width= 0.135\textwidth]{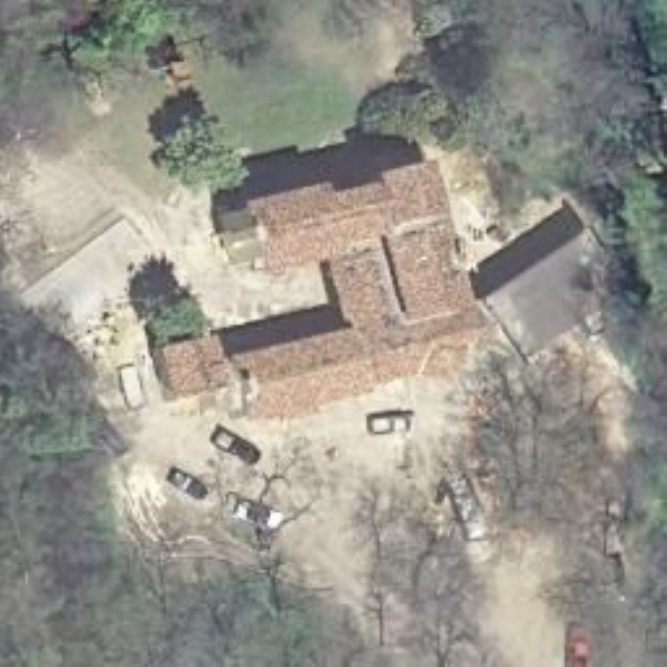}}
	\subfloat[]{
		\includegraphics[width= 0.135\textwidth]{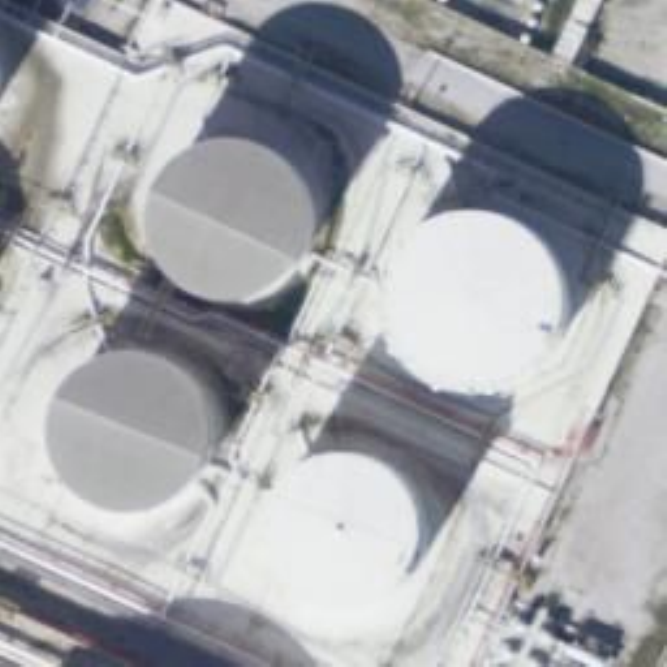}}
	\subfloat[]{
		\includegraphics[width= 0.135\textwidth]{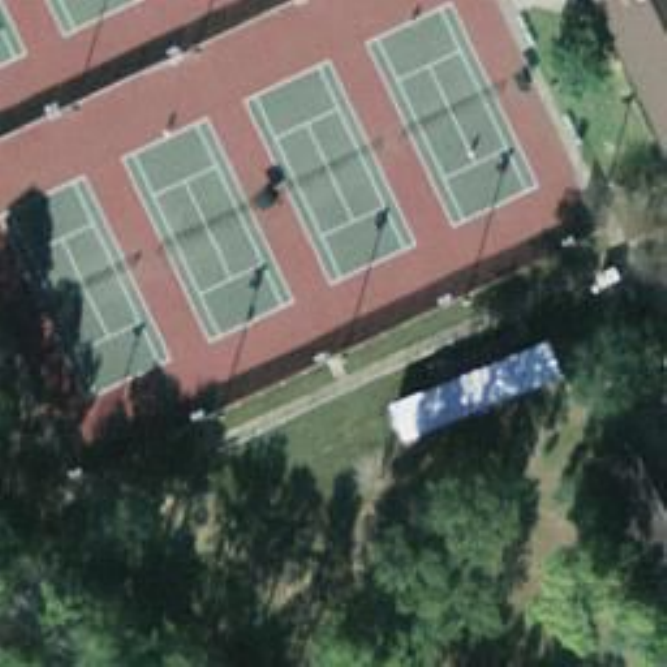}}
	\caption{Some examples from the UC Merced LULC data set. (1) agricultural. (2) airplane. (3) baseball. (4) beach. (5) buildings. (6) chaparral. (7) dense residential. (8) forest. (9) freeway. (10) golf course. (11) harbor. (12) intersection. (13) runway. (14) river. (15) medium residential. (16) parking lot. (17) overpass. (18) mobile home park. (19) sparse residential. (20) storage tanks. (21) tennis courts.}
	\label{uc}
	\vspace{-0.5em}
\end{figure*}

\section{Experiments}
\subsection{Implementation details}
For part1, we use the pre-trained ResNet~\cite{he2016deep} to produce attention maps consistent with the size of the original image. For part2, we experiment with the ResNet-18 architecture~\cite{he2016deep} and the SFT architecture we designed to generate 512x7x7-dim feature tensors. We use the ADAM optimizer~\cite{kingma2014adam} with AMSGrad~\cite{reddi2018convergence} by setting $\beta_1$ = 0.9, $\beta_2$ = 0.999 and $\epsilon$ = $10^{-8}$ to train the networks, and terminate after 40 epochs. We implement the proposed models via Pytorch framework and train them using NVIDIA K40 GPUs for acceleration.
\renewcommand{\thesubfigure}{\arabic{subfigure}}

\subsection{Data set Description}
UC-Merced dataset~\cite{yang2010bag} includes 2100 aerial remote sensing images of 21 classes, including agricultural, airplane, baseball diamond, beach, buildings, chaparral, dense residential, forest, freeway, golf course, harbor, intersection, medium density residential, mobile home park, overpass, parking lot, river, runway, sparse residential, storage tanks, and tennis courts. The size of the image, a pixel resolution of 0.3 m in the red green blue (RGB) color space, is 256 x 256 pixels. The dataset has a significant overlap among several classes, such as medium residential, dense residential and sparse residential, which make the dataset difficult for classification. Some examples are shown in Fig.~\ref{uc}.

\renewcommand{\thesubfigure}{\arabic{subfigure}}
\begin{figure*}[t]
	\centering
	\subfloat[]{
		\includegraphics[width= 0.1\textwidth]{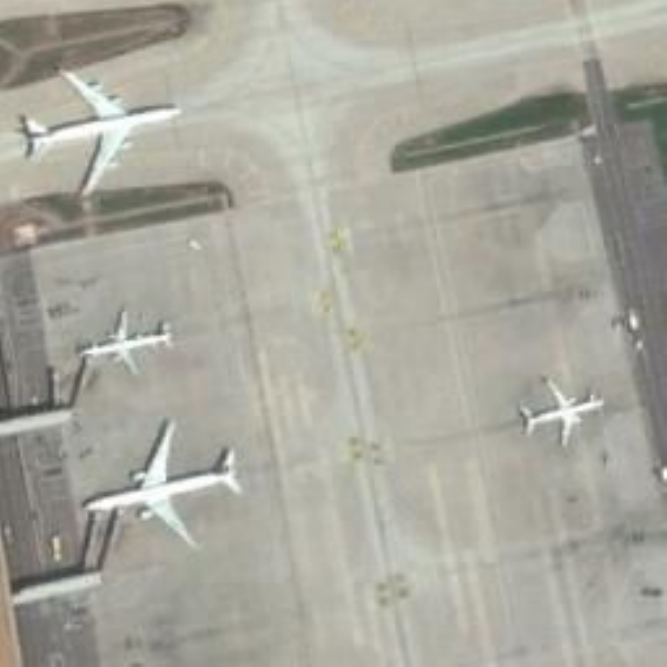}}
	\subfloat[]{
		\includegraphics[width= 0.1\textwidth]{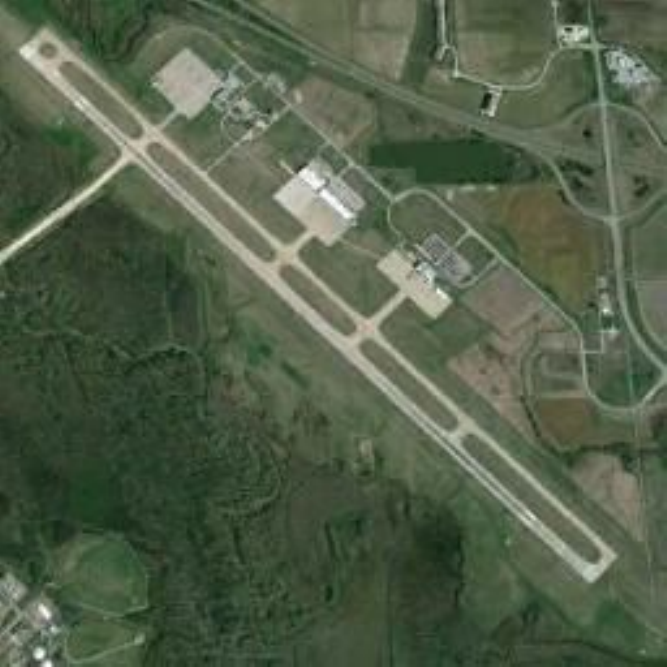}}
	\subfloat[]{
		\includegraphics[width= 0.1\textwidth]{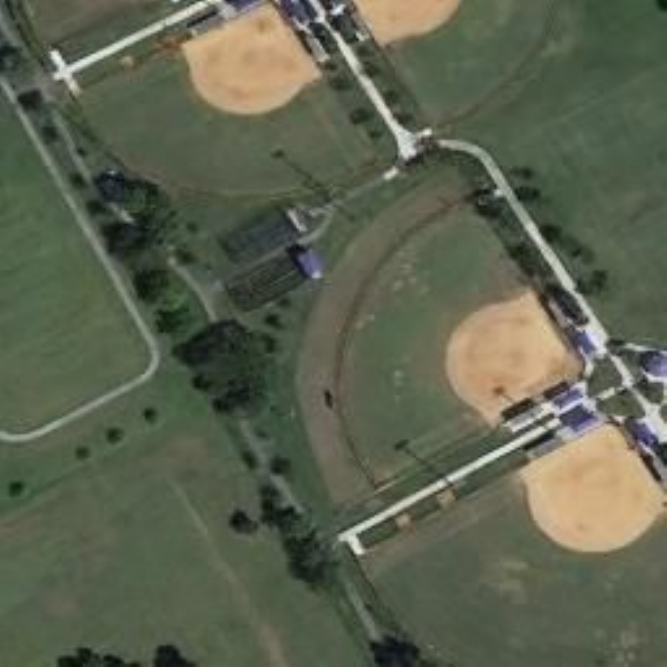}}
	\subfloat[]{
		\includegraphics[width= 0.1\textwidth]{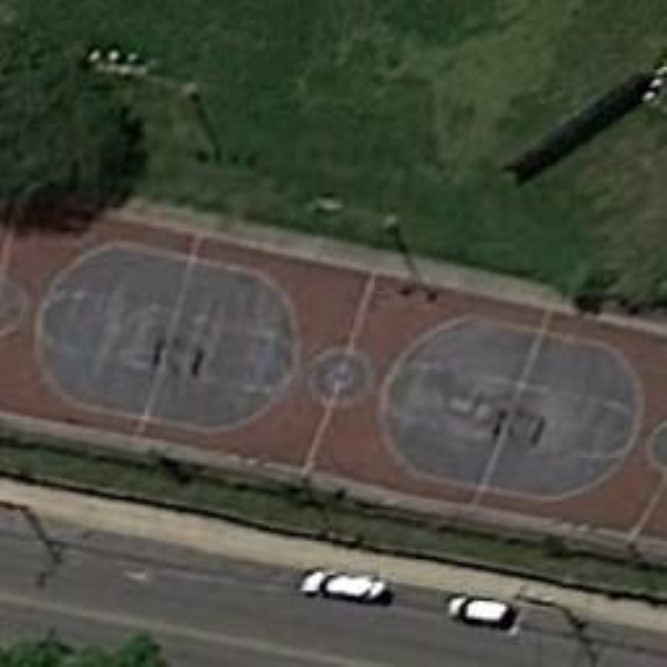}}
	\subfloat[]{
		\includegraphics[width= 0.1\textwidth]{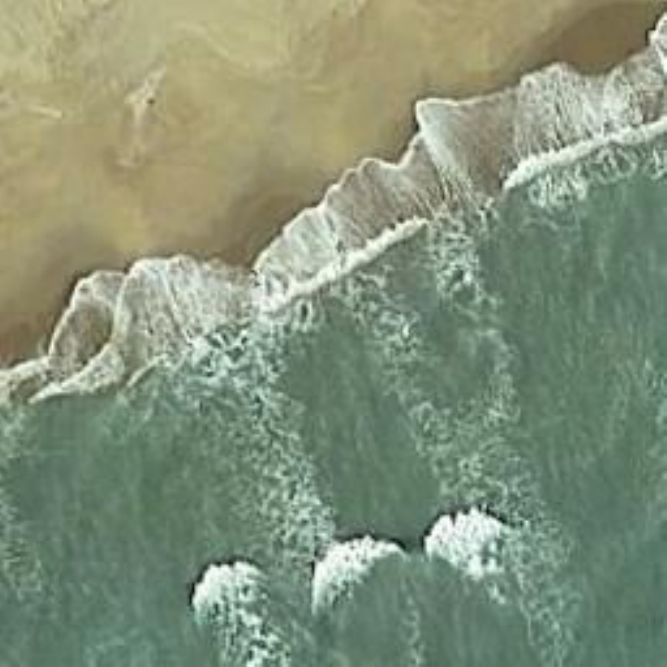}}
	\subfloat[]{
		\includegraphics[width= 0.1\textwidth]{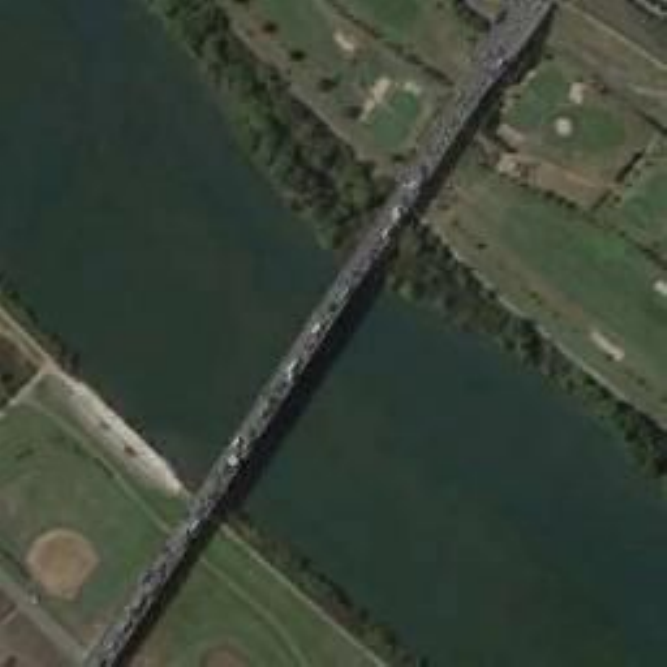}}
	\subfloat[]{
		\includegraphics[width= 0.1\textwidth]{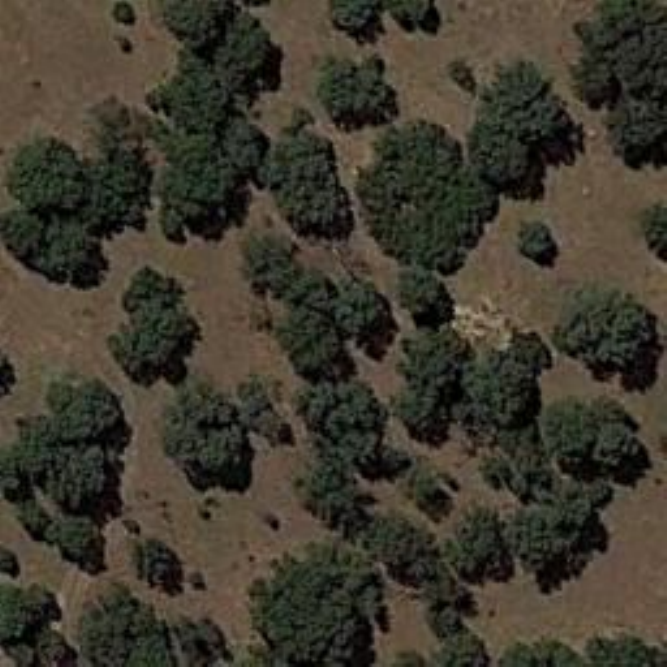}}
	\subfloat[]{
		\includegraphics[width= 0.1\textwidth]{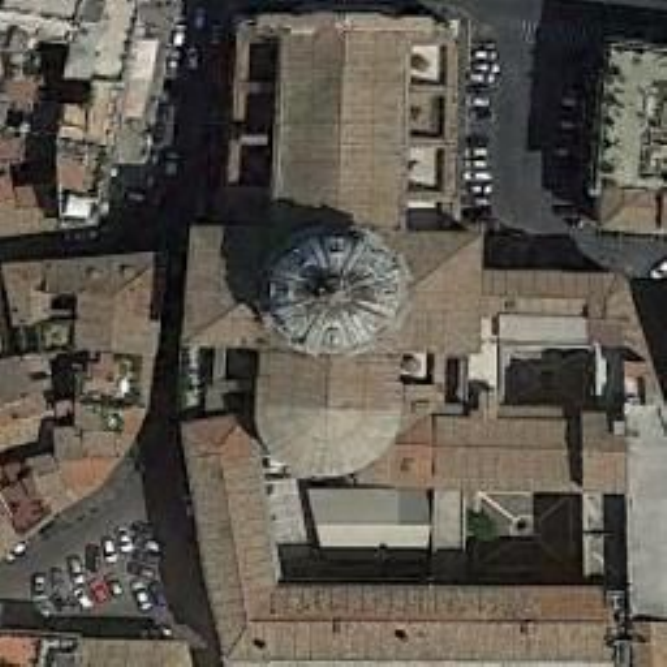}}
	\subfloat[]{
		\includegraphics[width= 0.1\textwidth]{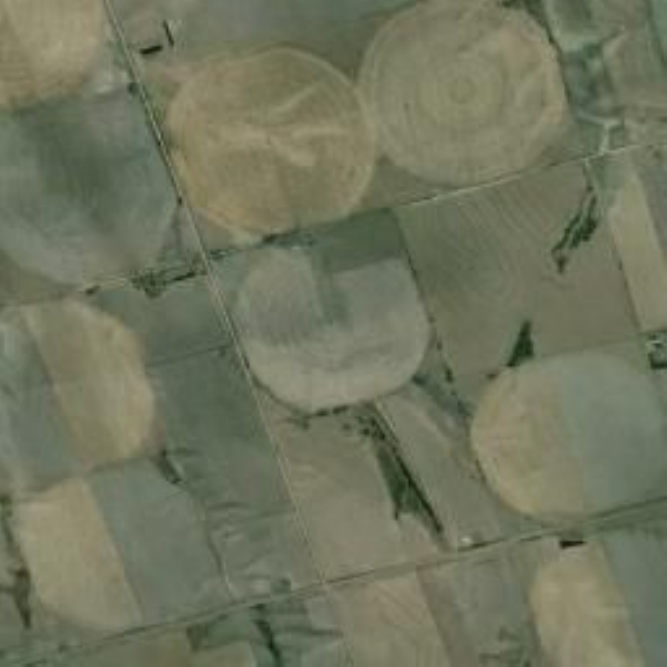}}
	\vspace{-0.5em}
	\hfill
	\subfloat[]{
		\includegraphics[width= 0.1\textwidth]{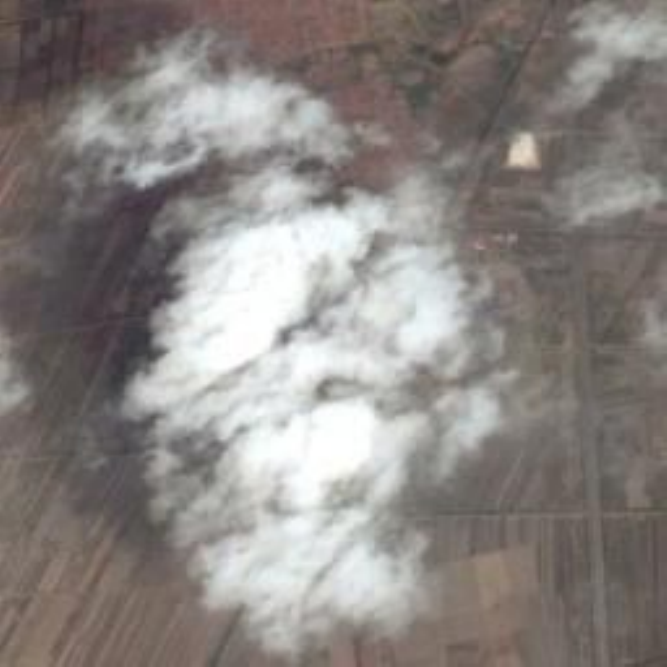}}
	\subfloat[]{
		\includegraphics[width= 0.1\textwidth]{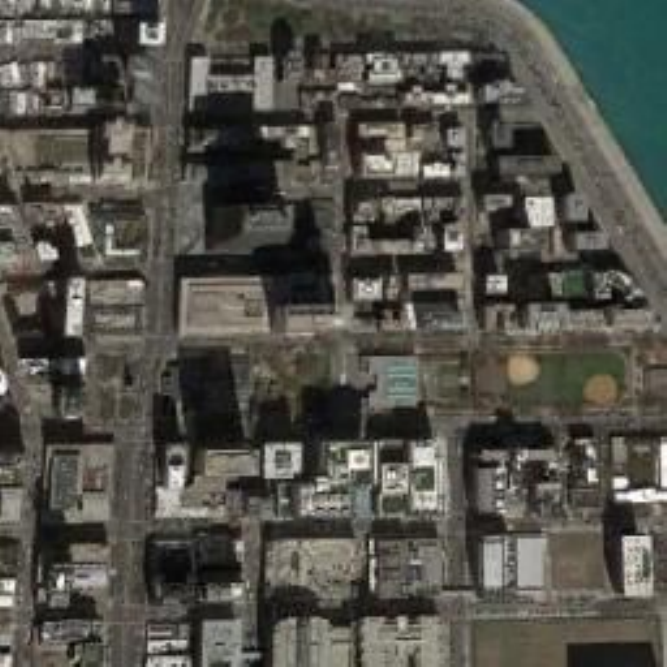}}
	\subfloat[]{
		\includegraphics[width= 0.1\textwidth]{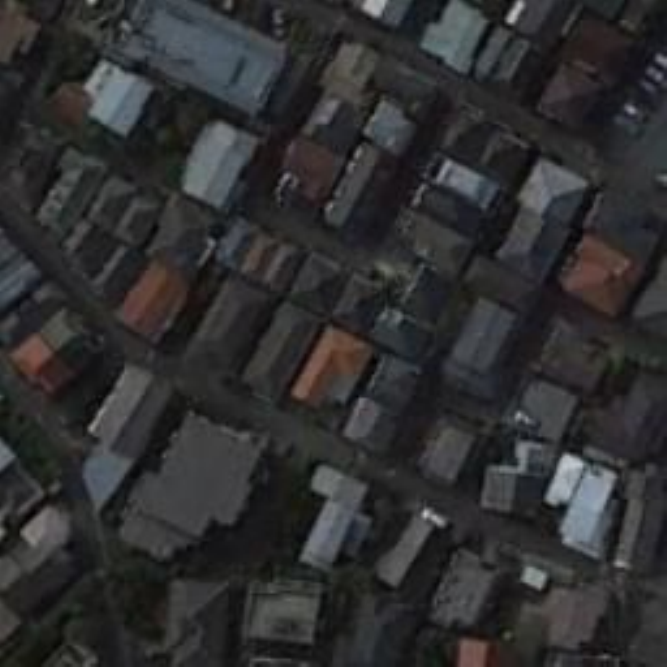}}
	\subfloat[]{
		\includegraphics[width= 0.1\textwidth]{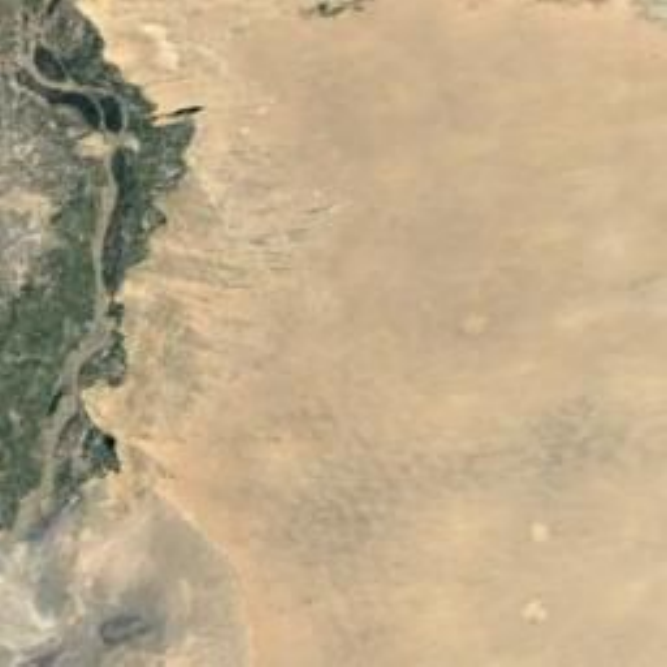}}
	\subfloat[]{
		\includegraphics[width= 0.1\textwidth]{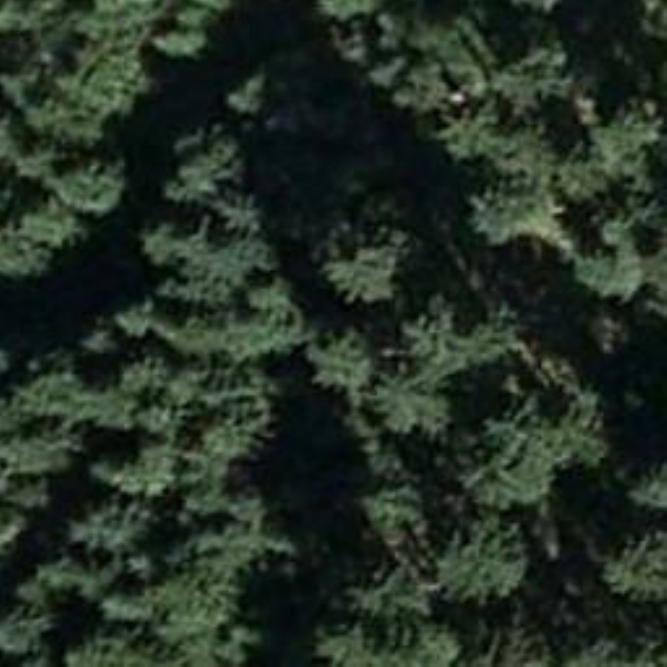}}
	\subfloat[]{
		\includegraphics[width= 0.1\textwidth]{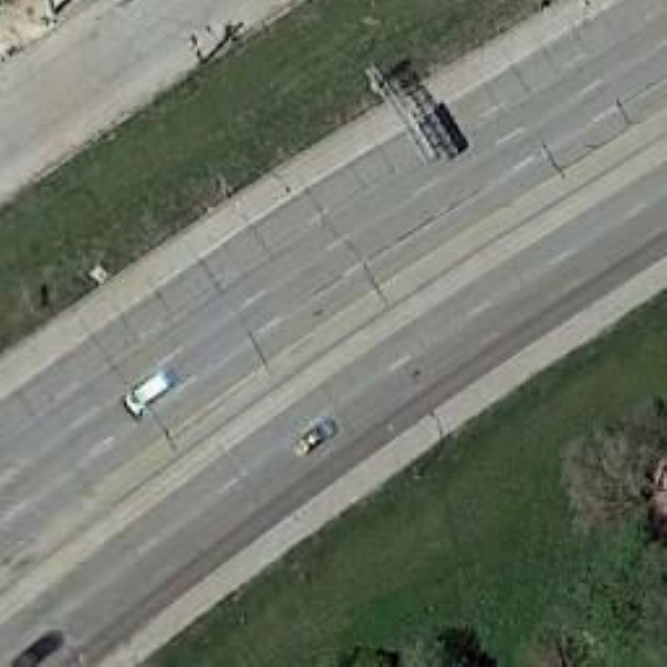}}
	\subfloat[]{
		\includegraphics[width= 0.1\textwidth]{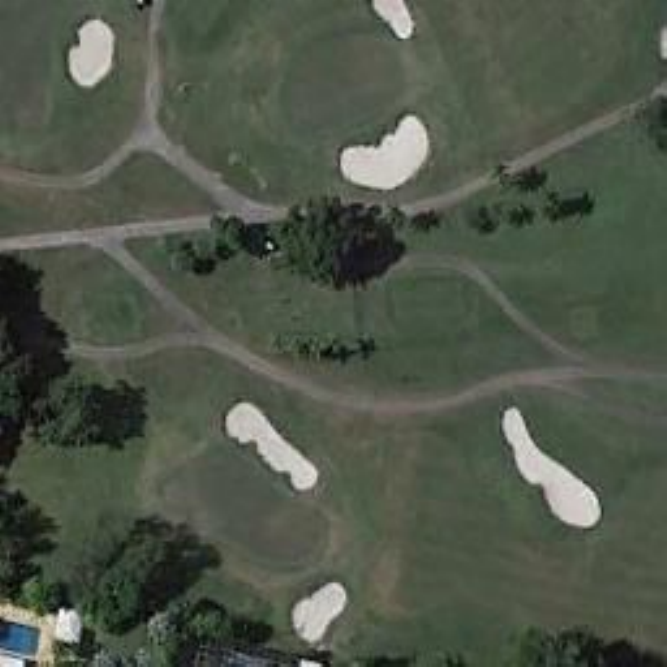}}
	\subfloat[]{
		\includegraphics[width= 0.1\textwidth]{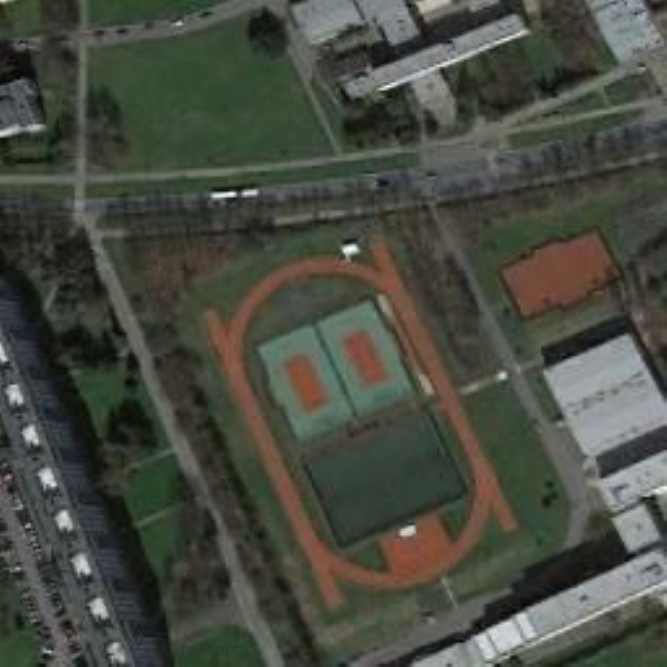}}
	\subfloat[]{
		\includegraphics[width= 0.1\textwidth]{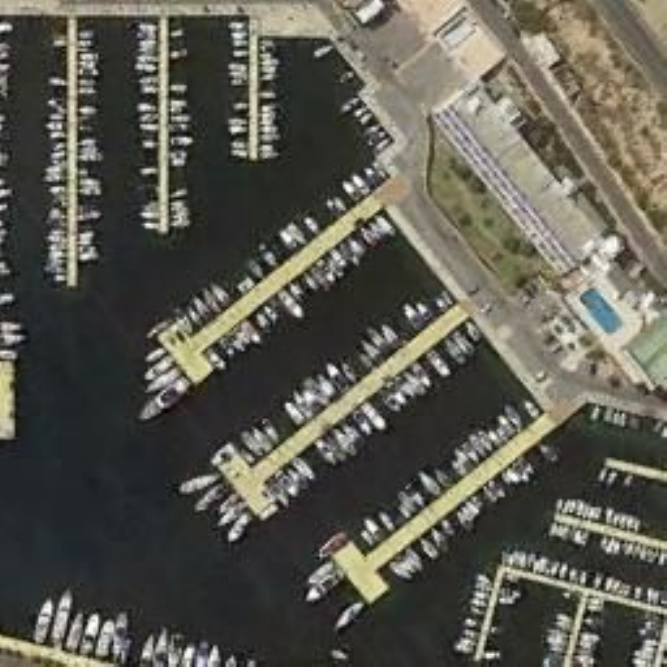}}
	\vspace{-0.5em}
	\hfill
	\subfloat[]{
		\includegraphics[width= 0.1\textwidth]{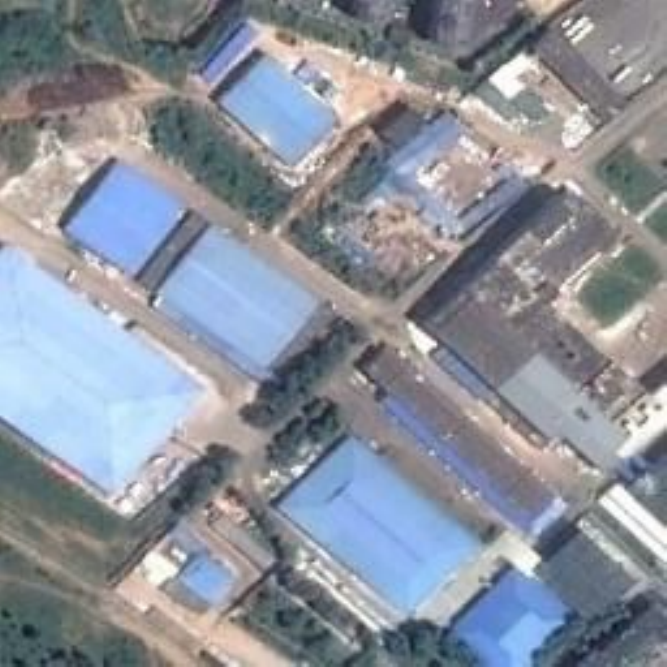}}
	\subfloat[]{
		\includegraphics[width= 0.1\textwidth]{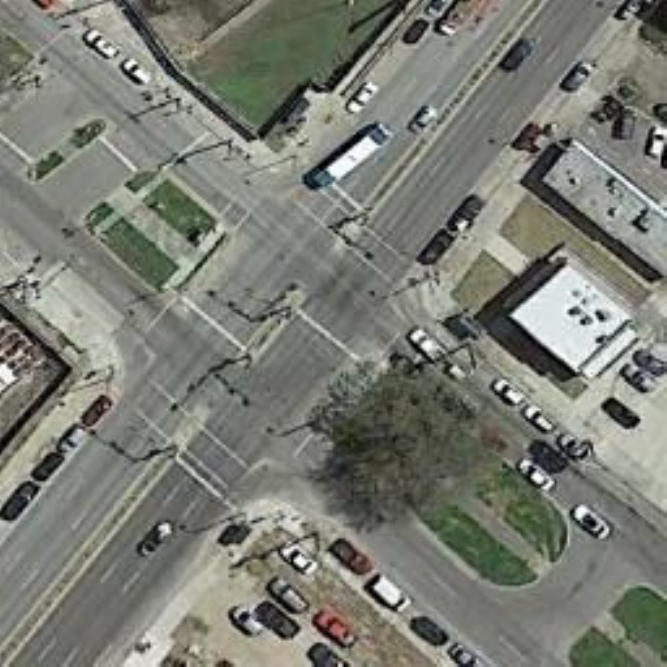}}
	\subfloat[]{
		\includegraphics[width= 0.1\textwidth]{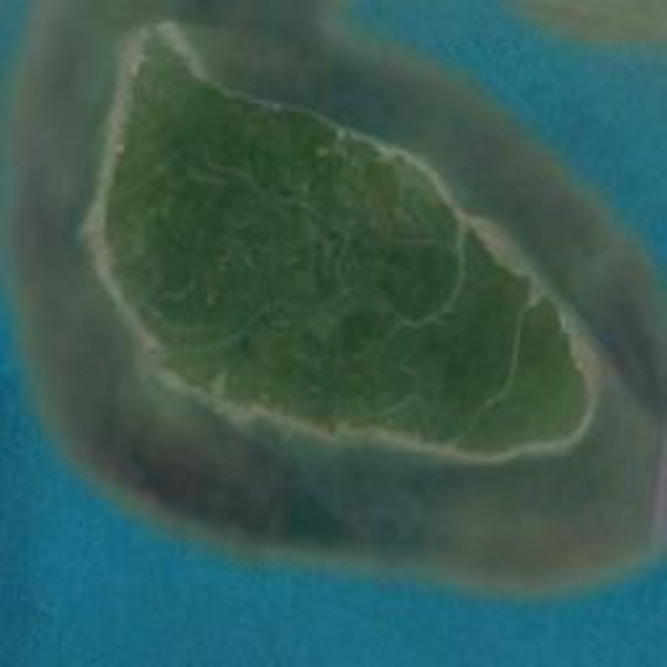}}	
	\subfloat[]{
		\includegraphics[width= 0.1\textwidth]{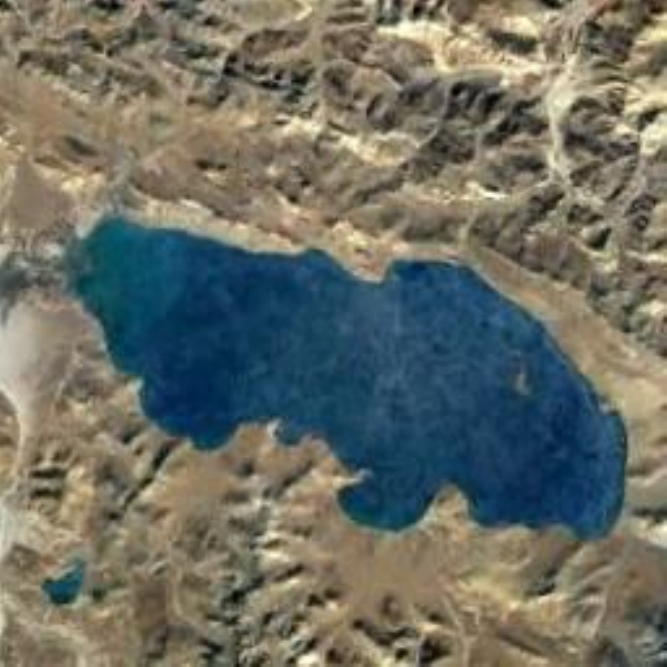}}
	\subfloat[]{
		\includegraphics[width= 0.1\textwidth]{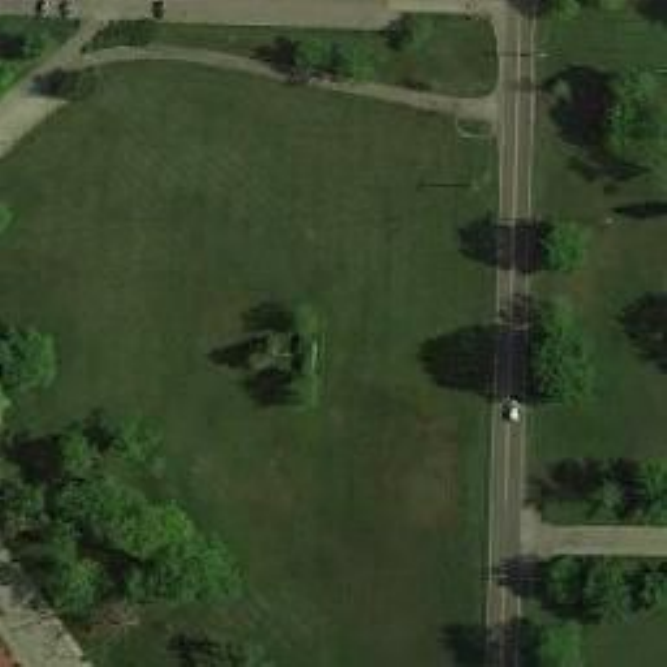}}
	\subfloat[]{
		\includegraphics[width=0.1\textwidth]{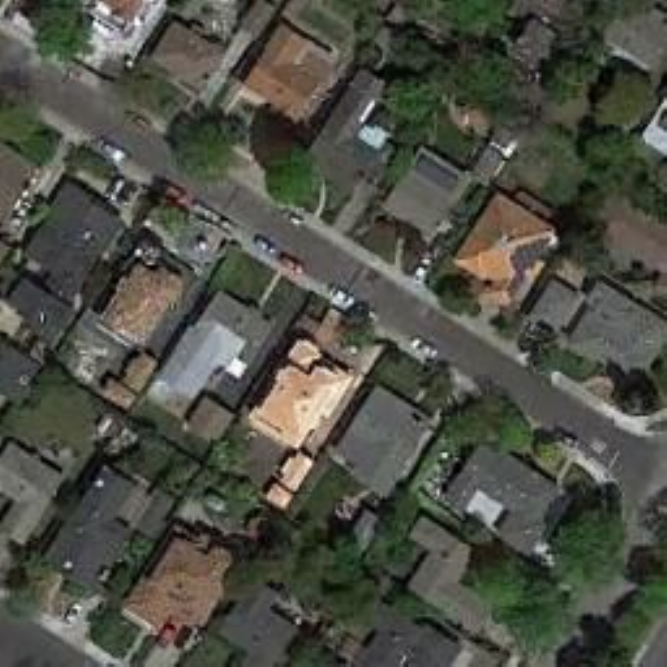}}
	\subfloat[]{
		\includegraphics[width=0.1\textwidth]{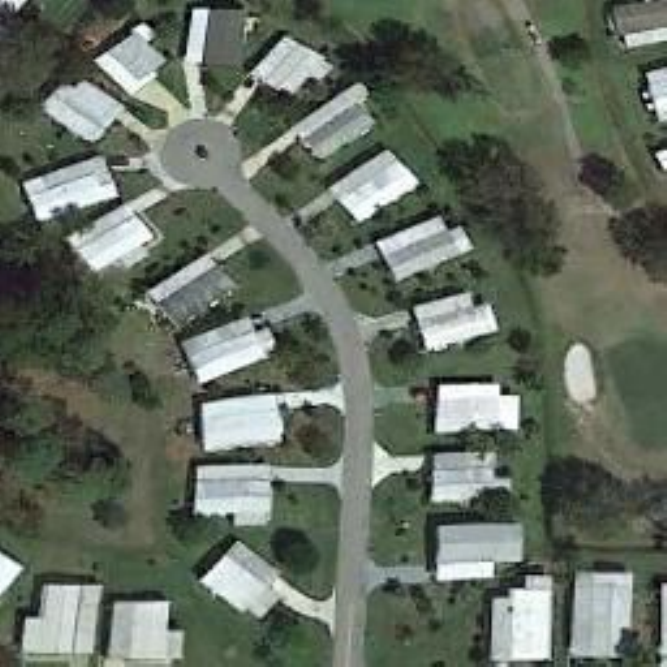}}
	\subfloat[]{
		\includegraphics[width= 0.1\textwidth]{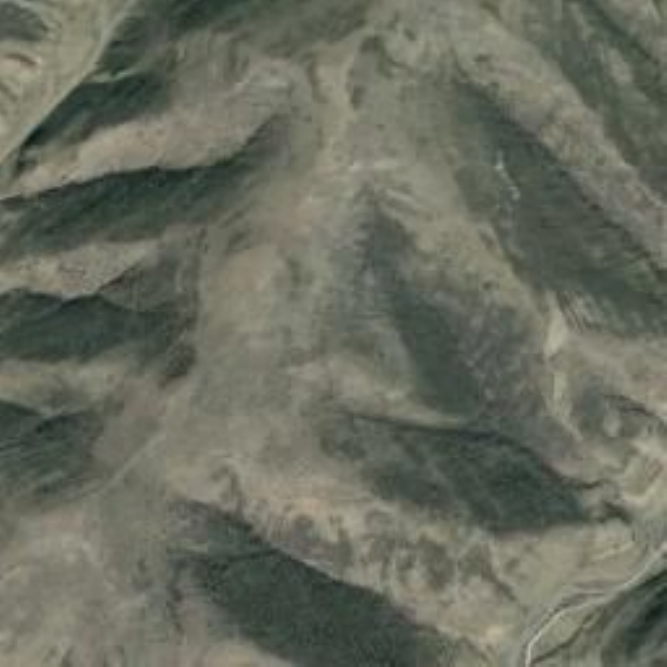}}
	\subfloat[]{
		\includegraphics[width= 0.1\textwidth]{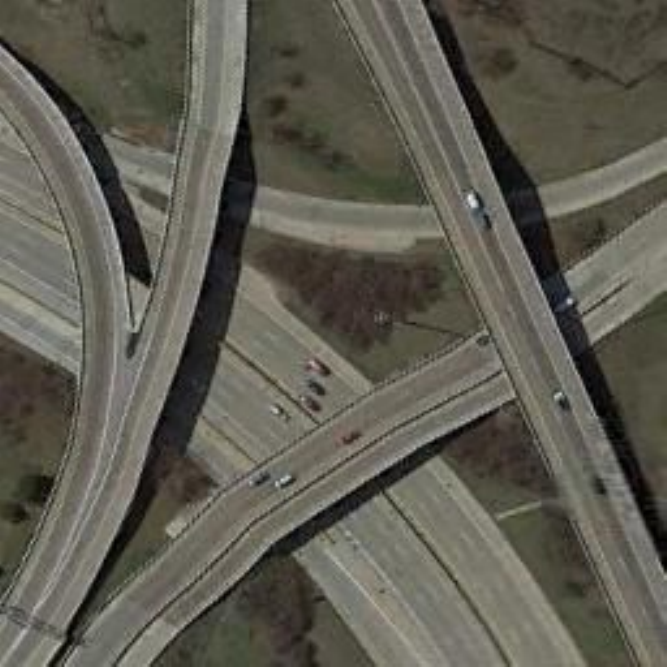}}
	\vspace{-0.5em}
	\hfill
	\subfloat[]{
		\includegraphics[width= 0.1\textwidth]{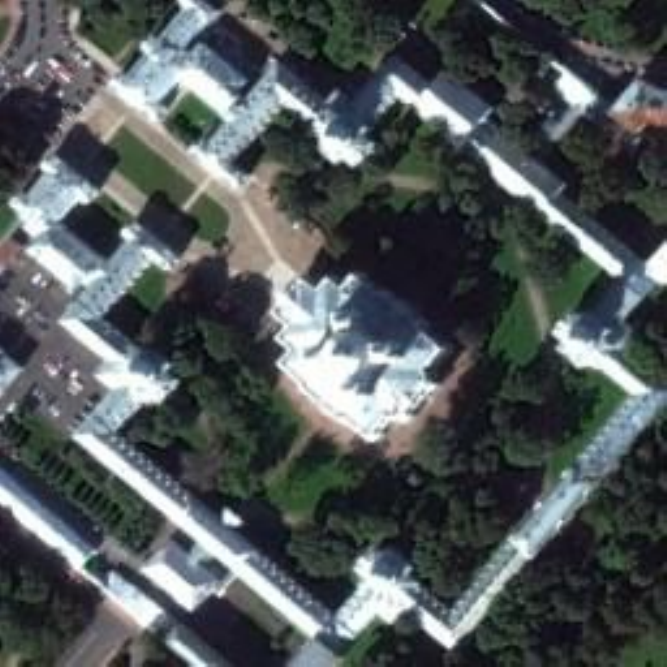}}
	\subfloat[]{
		\includegraphics[width= 0.1\textwidth]{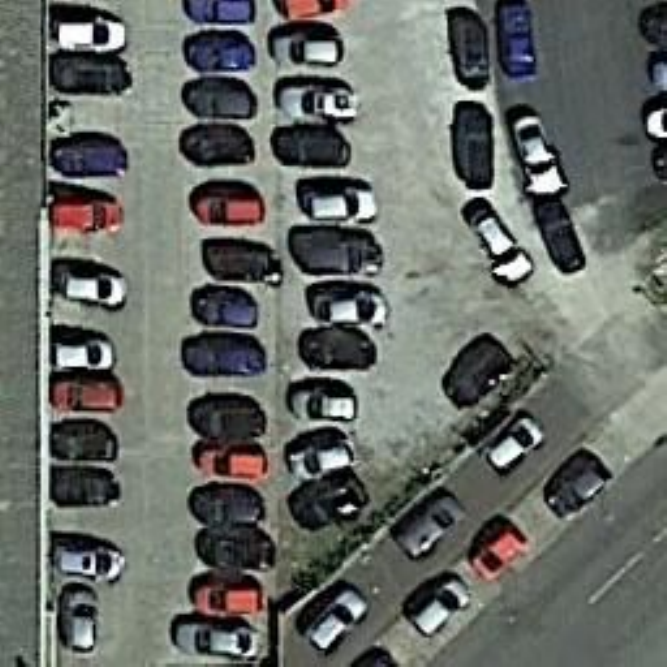}}
	\subfloat[]{
		\includegraphics[width= 0.1\textwidth]{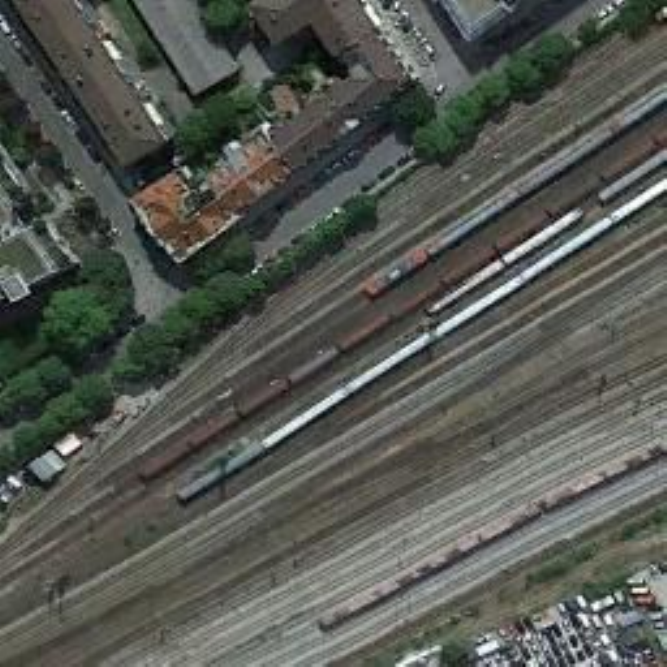}}
	\subfloat[]{
		\includegraphics[width= 0.1\textwidth]{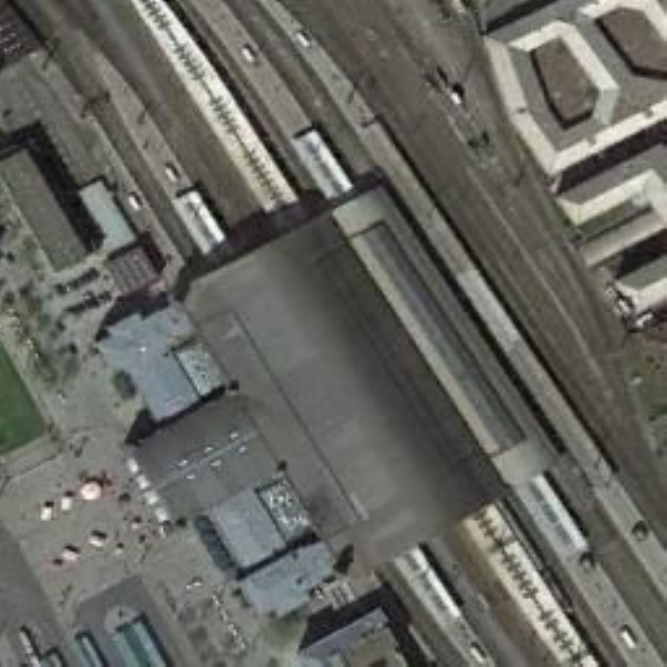}}
	\subfloat[]{
		\includegraphics[width=0.1\textwidth]{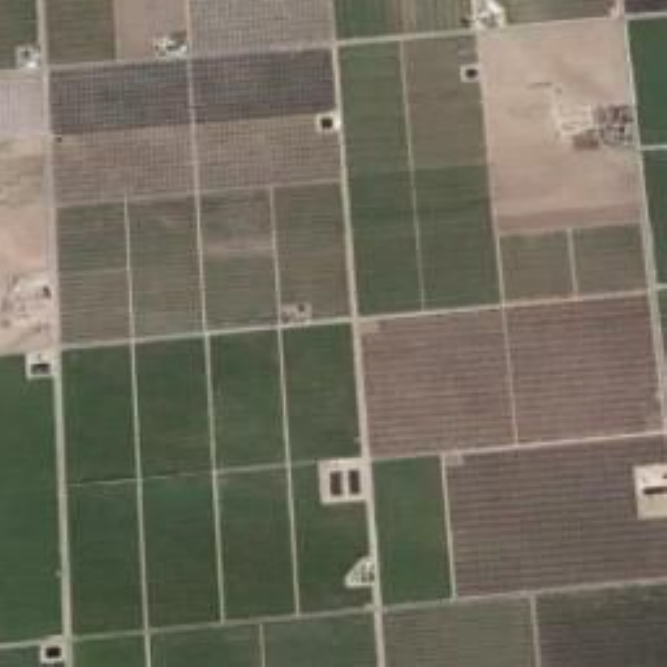}}
	\subfloat[]{
		\includegraphics[width= 0.1\textwidth]{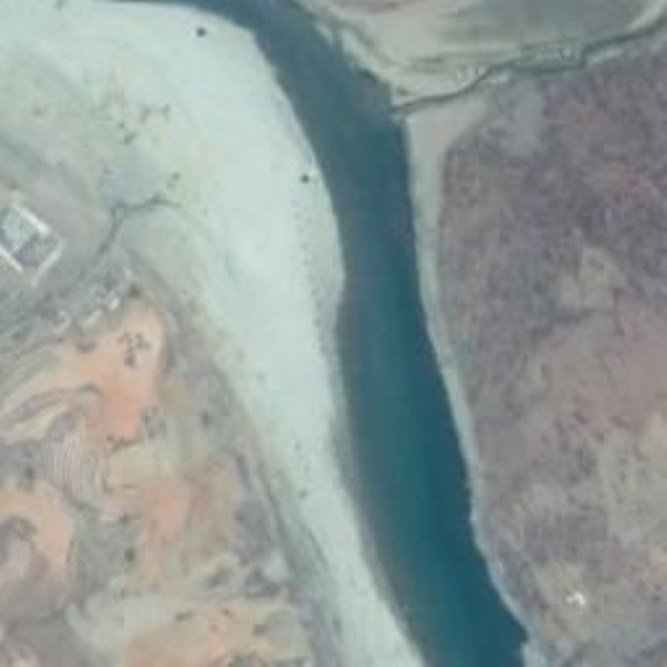}}
	\subfloat[]{
		\includegraphics[width= 0.1\textwidth]{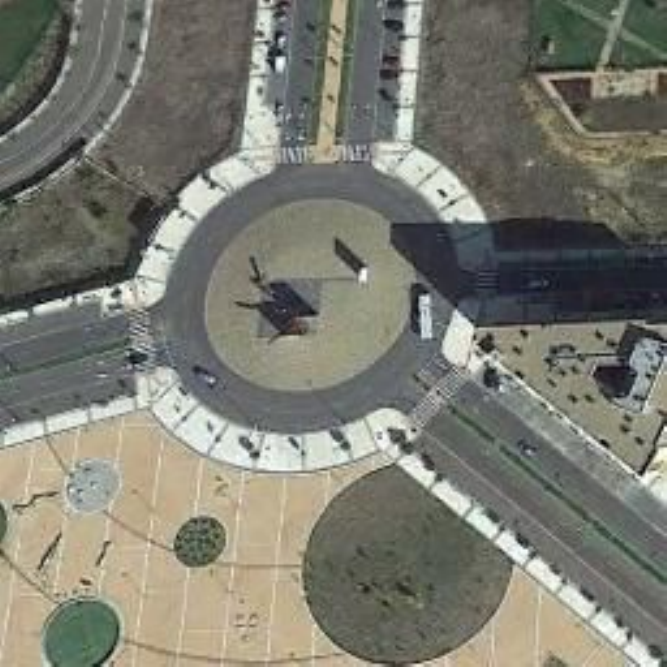}}
	\subfloat[]{
		\includegraphics[width= 0.1\textwidth]{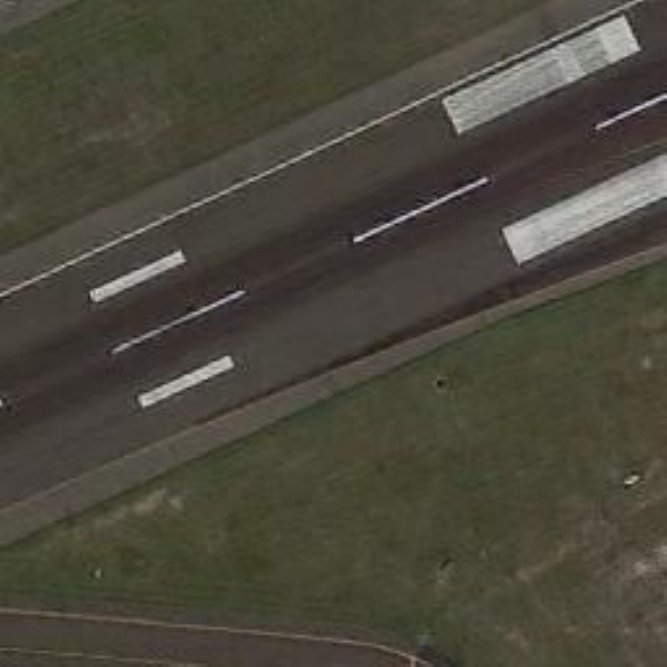}}
	\subfloat[]{
		\includegraphics[width= 0.1\textwidth]{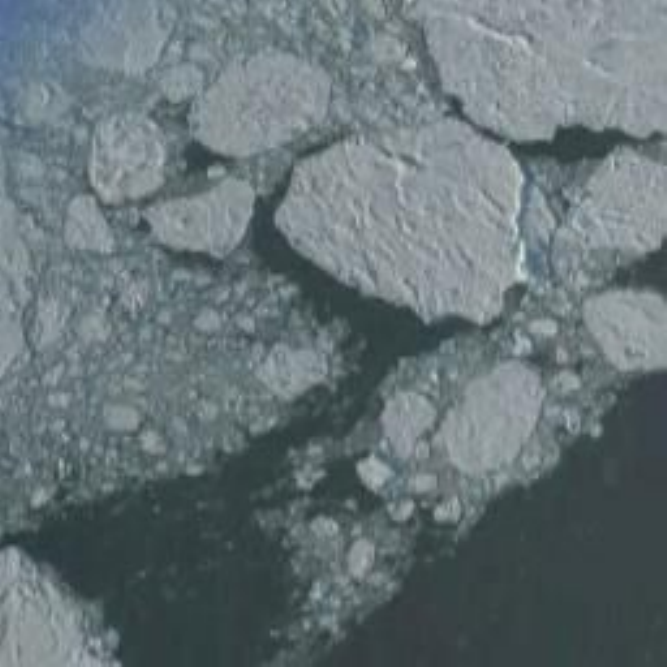}}
	\vspace{-0.5em}
	\hfill
	\subfloat[]{
		\includegraphics[width= 0.1\textwidth]{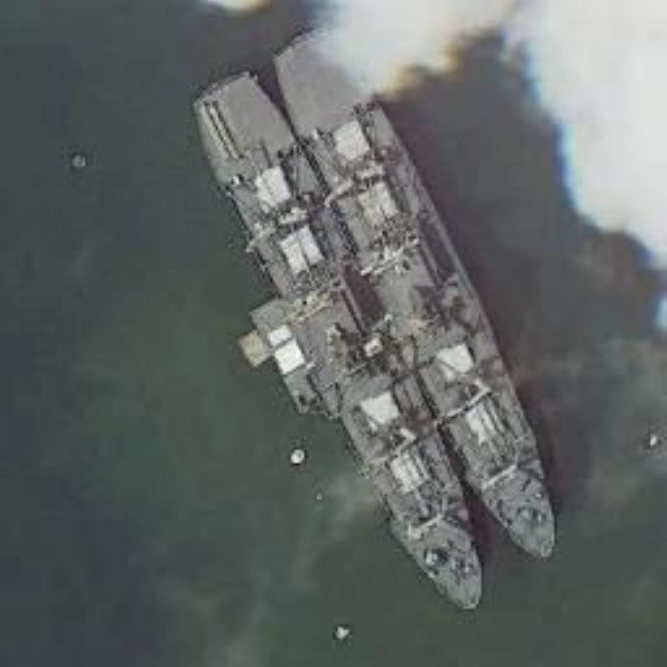}}
	\subfloat[]{
		\includegraphics[width= 0.1\textwidth]{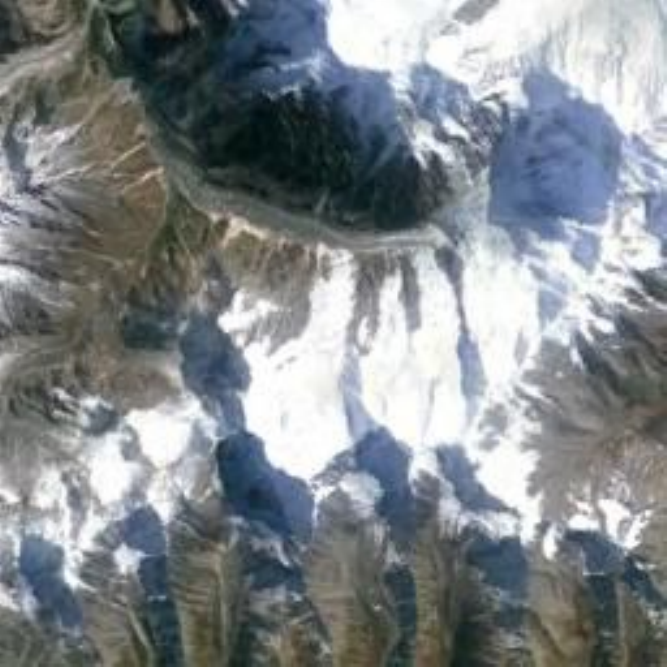}}
	\subfloat[]{
		\includegraphics[width=0.1\textwidth]{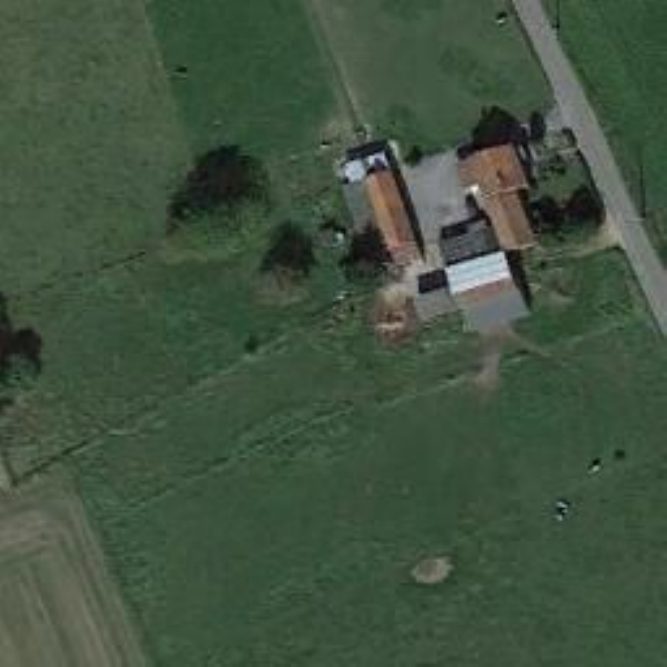}}
	\subfloat[]{
		\includegraphics[width= 0.1\textwidth]{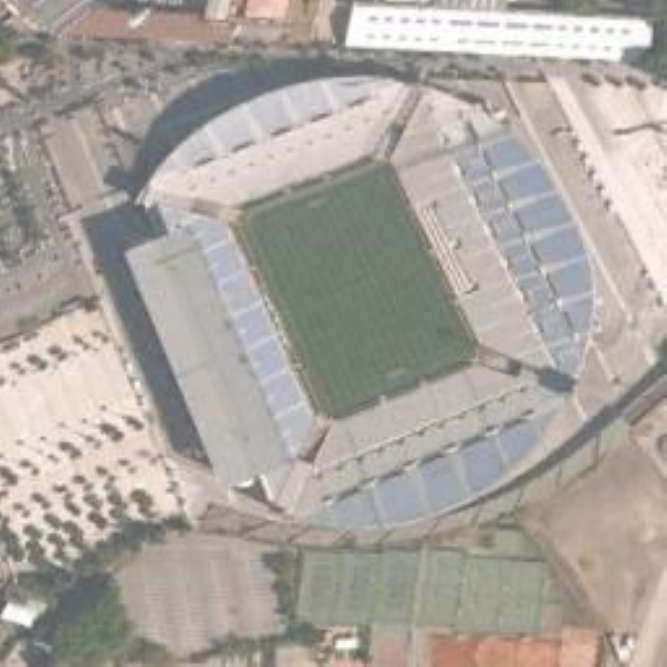}}
	\subfloat[]{
		\includegraphics[width= 0.1\textwidth]{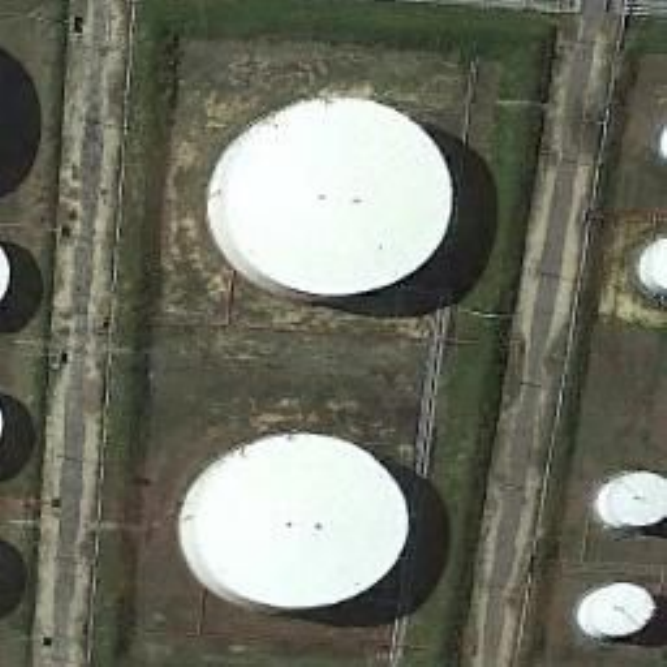}}
	\subfloat[]{
		\includegraphics[width= 0.1\textwidth]{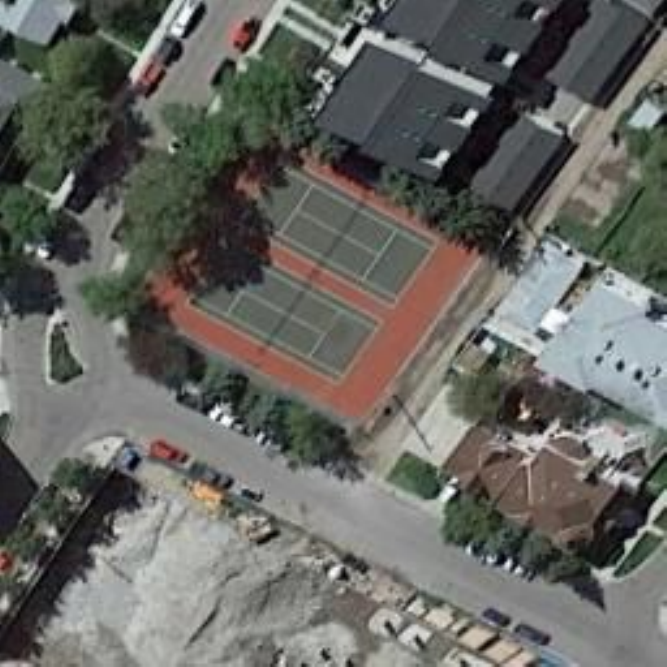}}
	\subfloat[]{
		\includegraphics[width= 0.1\textwidth]{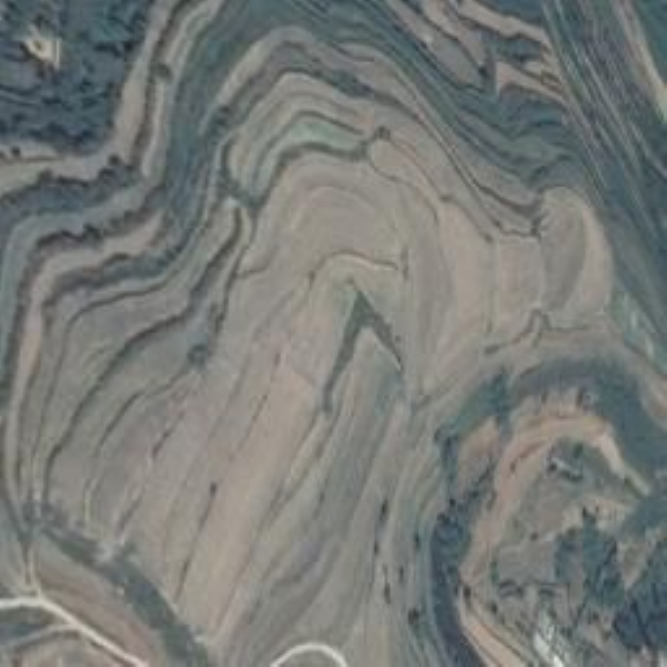}}
	\subfloat[]{
		\includegraphics[width=0.1\textwidth]{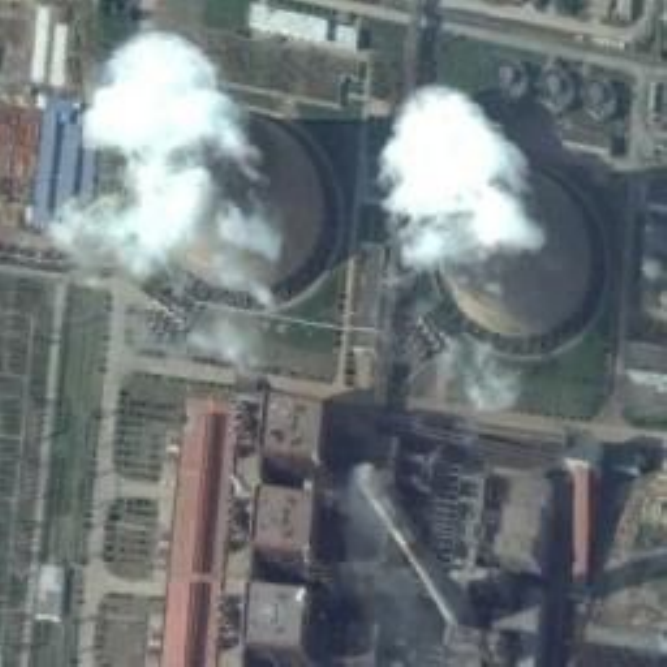}}
	\subfloat[]{
		\includegraphics[width= 0.1\textwidth]{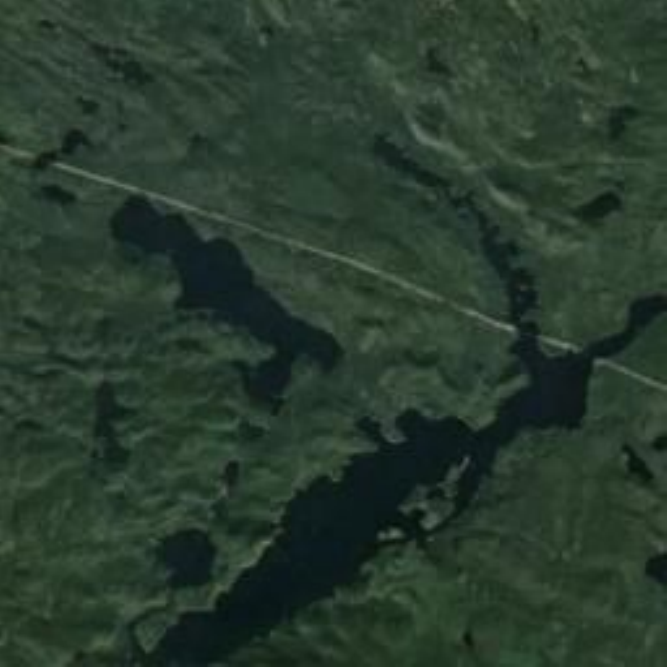}}
	\vspace{-0.5em}
	\caption{Example images from the NWPU-RESISC45 data set. (1) Airplane. (2) Airport. (3) Baseball diamond. (4) Basketball court. (5) Beach. (6) Bridge. (7) Chaparral. (8) Church. (9) Circular farmland. (10) Cloud. (11) Commercial area. (12) Dense residential. (13) Desert. (14) Forest. (15) Freeway. (16) Golf course. (17) Ground track field. (18) Harbor. (19) Industrial area. (20) Intersection. (21) Island. (22) Lake. (23) Meadow. (24) Medium residential. (25) Mobile home park. (26) Mountain. (27) Overpass. (28) Palace. (29) Parking lot. (30) Railway. (31) Railway station. (32) Rectangular farmland. (33) River. (34) Roundabout. (35) Runway. (36) Sea ice. (37) Ship. (38) Snowberg. (39) Sparse residential. (40) Stadium. (41) Storage tank. (42) Tennis court. (43) Terrace. (44) Thermal power station. (45) Wetland}
	\label{nwpu45}
\end{figure*}

\begin{figure*}[t]
	\centering
	\subfloat{
		\includegraphics[width= 0.16\textwidth]{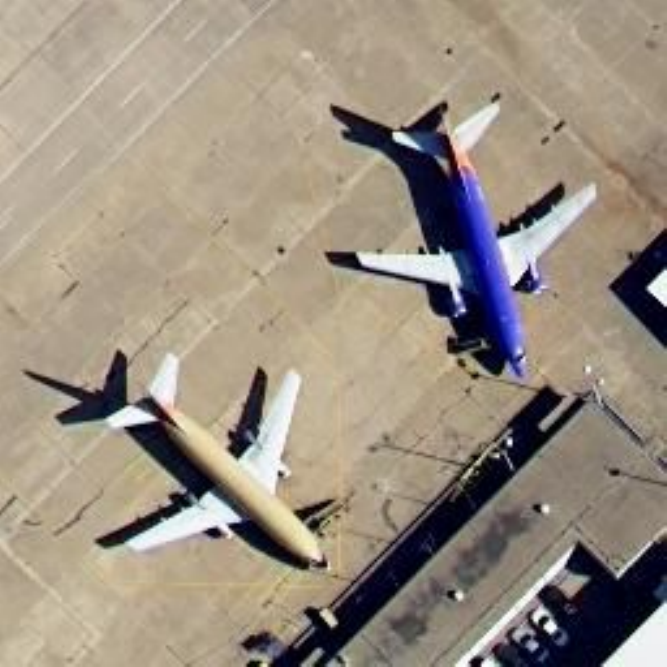}}
	\subfloat{
		\includegraphics[width= 0.16\textwidth]{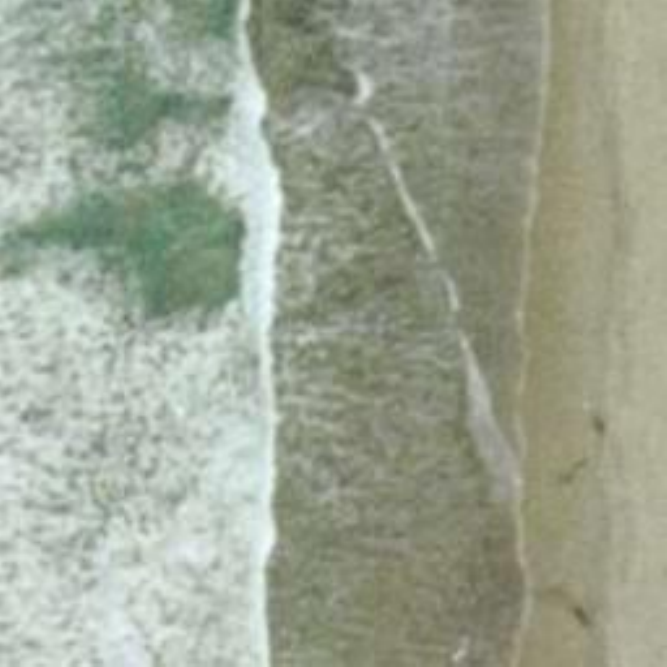}}
	\subfloat{
		\includegraphics[width= 0.16\textwidth]{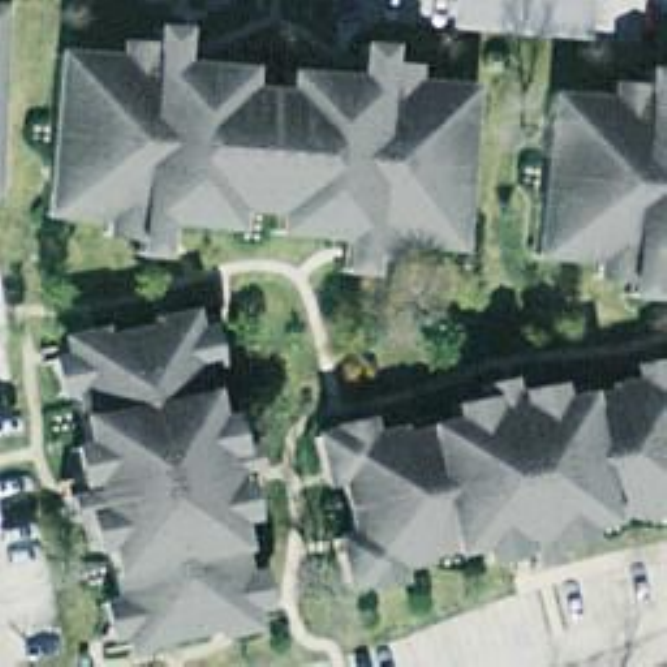}}
	\subfloat{
		\includegraphics[width= 0.16\textwidth]{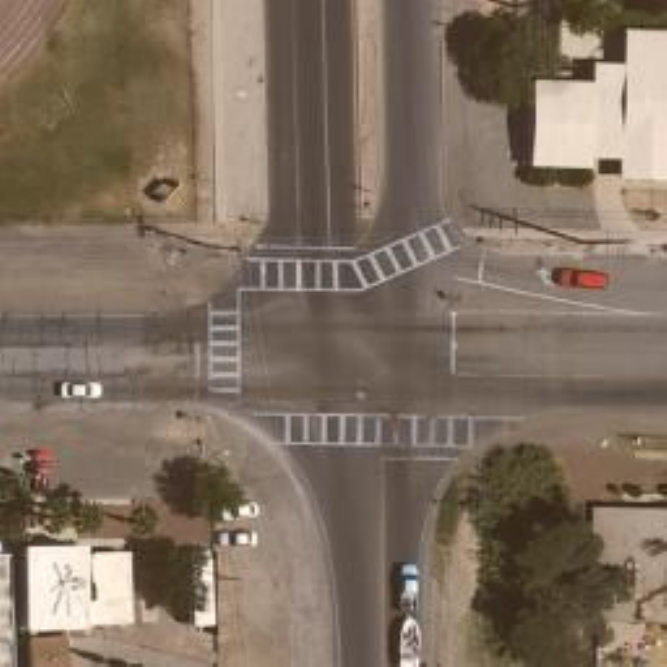}}
	\subfloat{
		\includegraphics[width= 0.16\textwidth]{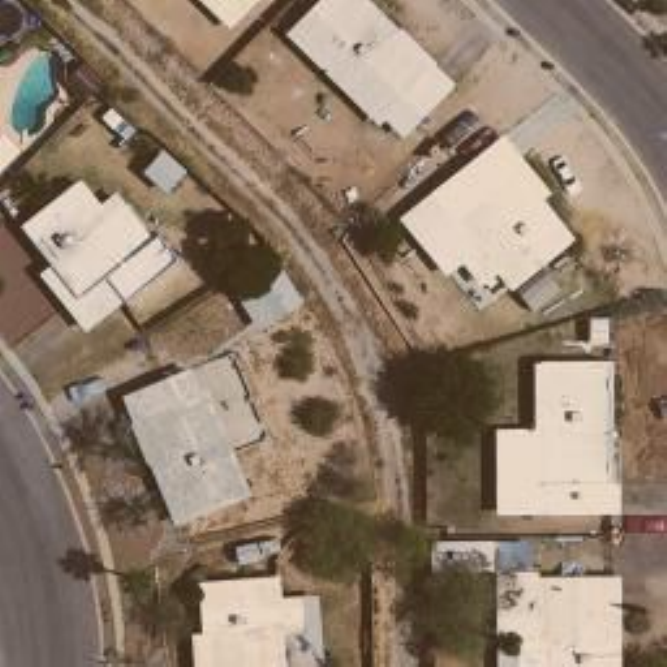}}
	\subfloat{
		\includegraphics[width= 0.16\textwidth]{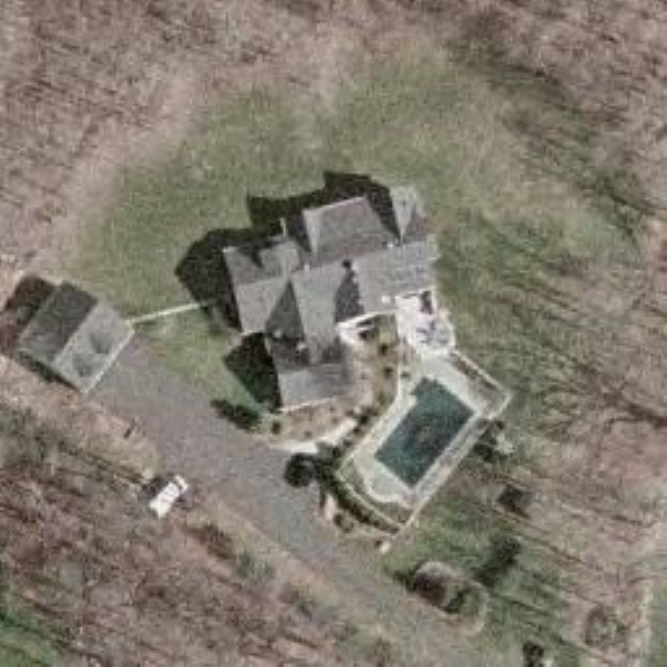}}
	\vspace{-0.5em}
	\hfill
	\subfloat{
		\includegraphics[width= 0.16\textwidth]{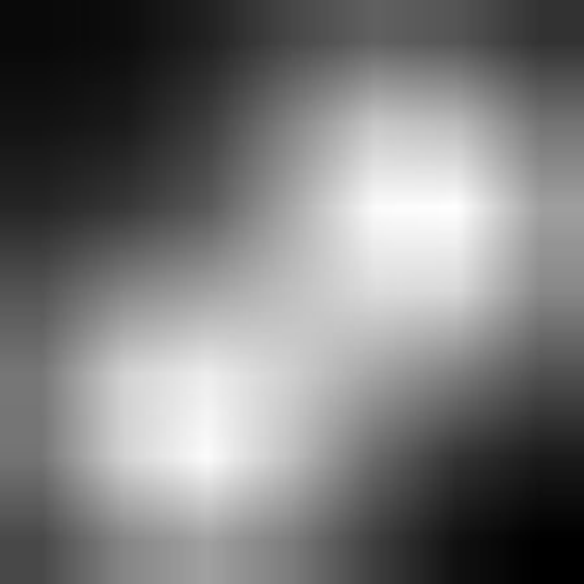}}
	\subfloat{
		\includegraphics[width= 0.16\textwidth]{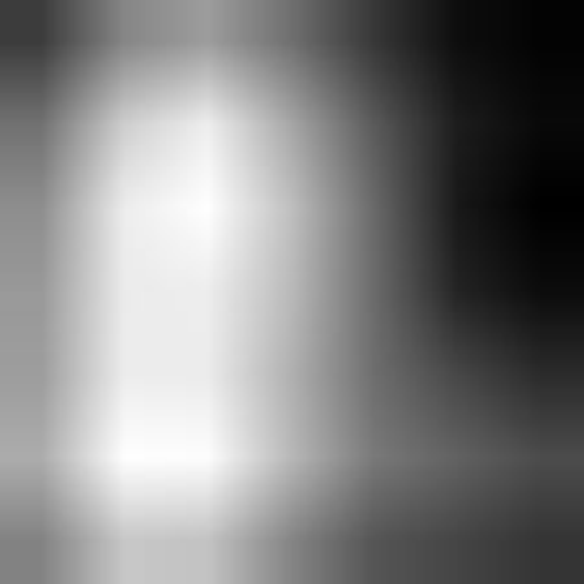}}
	\subfloat{
		\includegraphics[width= 0.16\textwidth]{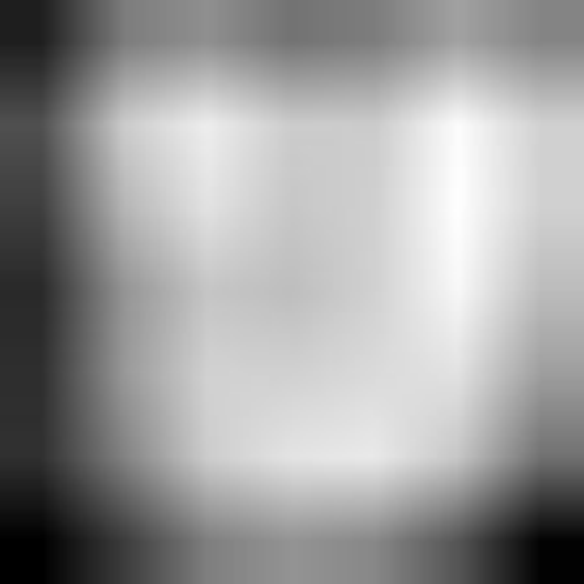}}
	\subfloat{
		\includegraphics[width= 0.16\textwidth]{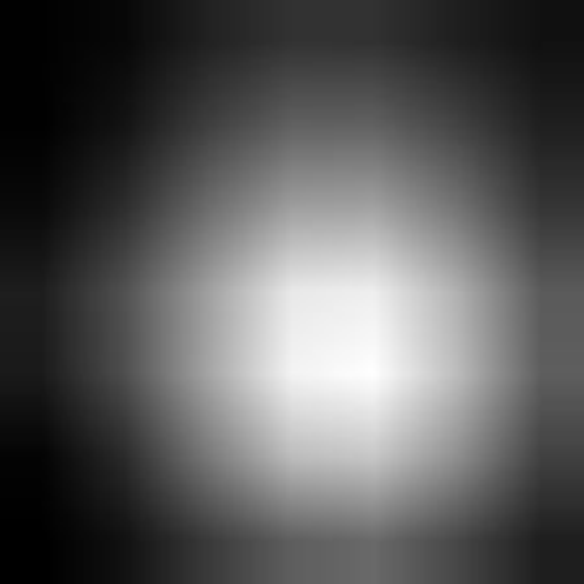}}
	\subfloat{
		\includegraphics[width= 0.16\textwidth]{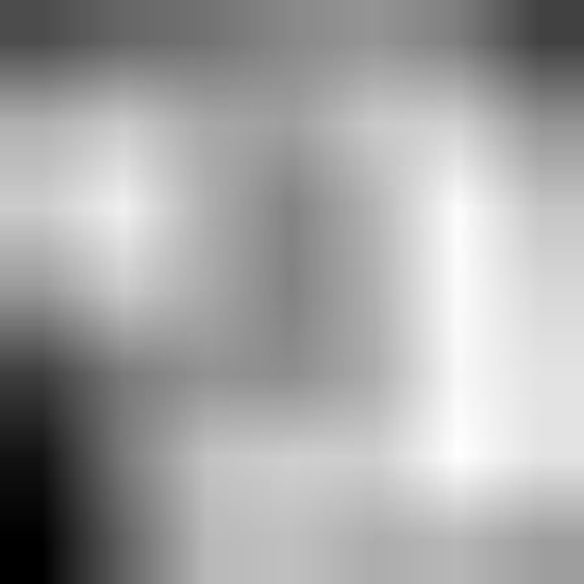}}
	\subfloat{
		\includegraphics[width= 0.16\textwidth]{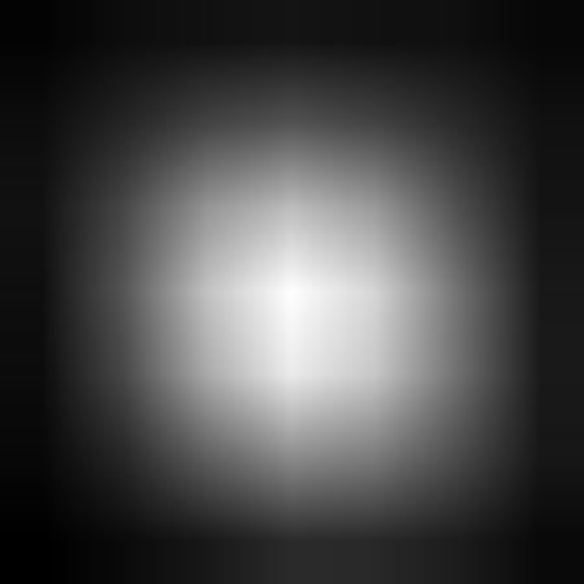}}
		\vspace{-0.5em}
	\hfill
	\subfloat{
		\includegraphics[width= 0.16\textwidth]{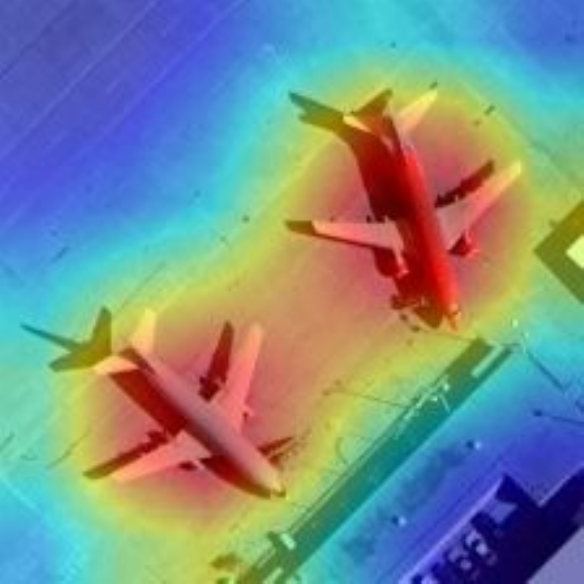}}
	\subfloat{
		\includegraphics[width= 0.16\textwidth]{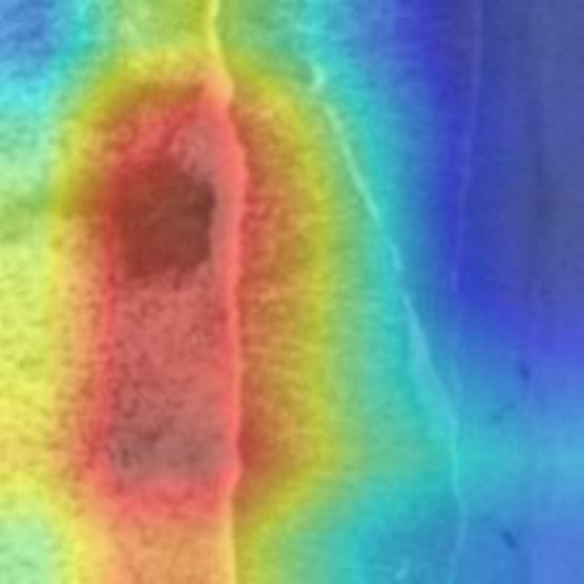}}
	\subfloat{
		\includegraphics[width= 0.16\textwidth]{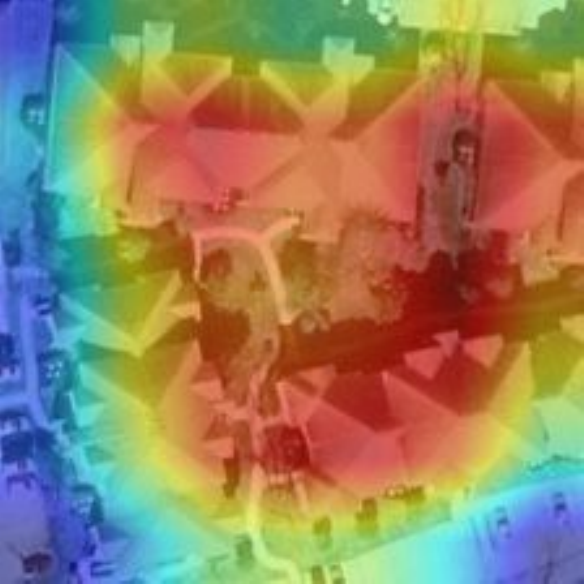}}
	\subfloat{
		\includegraphics[width= 0.16\textwidth]{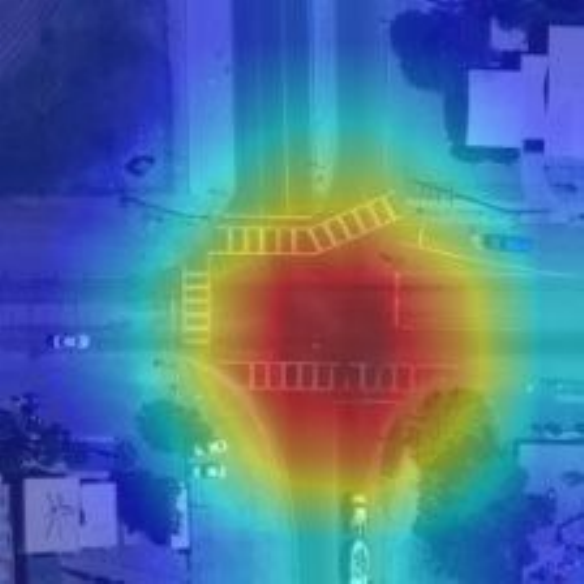}}
	\subfloat{
		\includegraphics[width= 0.16\textwidth]{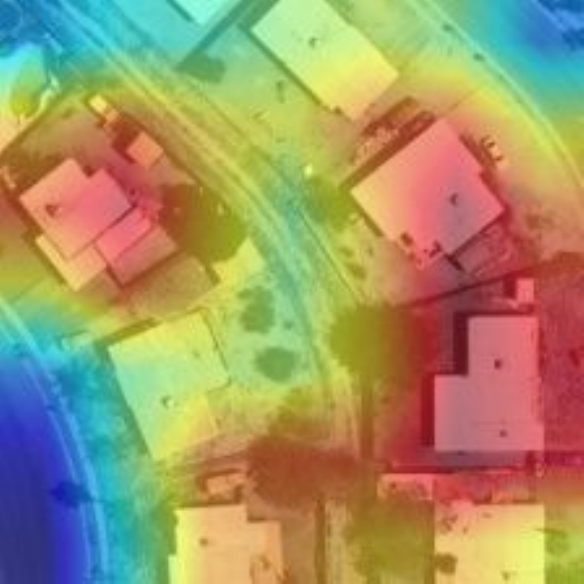}}
	\subfloat{
		\includegraphics[width= 0.16\textwidth]{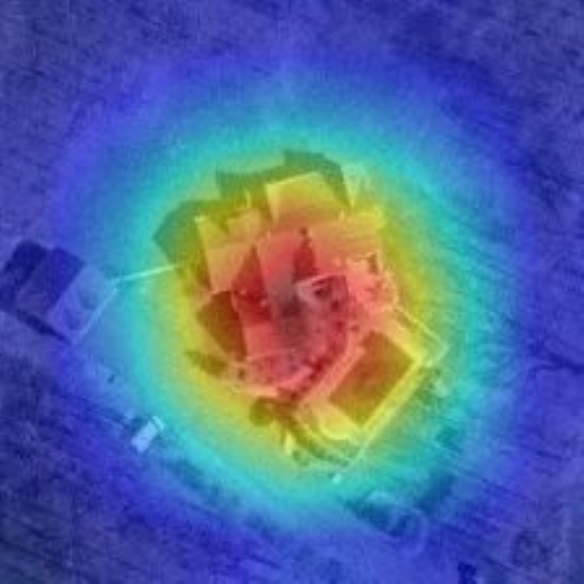}}
	\caption{Saliency Mask results. The grad-CAM visualization is calculated for the last convolutional outputs. The ground-truth label is shown on the top of each input image (first line) and predicted-label is displayed on the top of saliency image (second line and third line) }
	\label{atten}
	\vspace{-0.5em}
\end{figure*}

The NWPU-RESISC45 dataset~\cite{cheng2017remote} includes totally 31500 remote sensing images divided into 45 scene classes. Each class consists of 700 images with a size of 256 x 256 pixels in the RGB color space. The range of spatial resolution is from about 30 to 0.2 m per pixel for most of the scene classes.  This dataset contains more than 100 urban areas around the world. These 45 scene classes are airplane, airport, baseball diamond, basketball court, beach, bridge, chaparral, church, circular farmland, cloud, commercial area, dense residential, desert, forest, freeway, golf course, ground track field, harbor, industrial area, intersection, island, lake, meadow, medium residential, mobile home park, mountain, overpass, palace, parking lot, railway, railway station, rectangular farmland, river, roundabout, runway, sea ice, ship, snow berg, sparse residential, stadium, storage tank, tennis court, terrace, thermal power station, and wetland. The high within class diversity and between-class similarity make the data set more challenging. Some examples are shown in Fig.~\ref{nwpu45}.

\subsection{Experimental Setting}
\subsubsection{Data Set Setting and Evaluation Metrics}

For the UC Merced data set~\cite{yang2010bag}, we randomly split it into 80\% for training and 20\% for testing. For the NWPU-RESISC45 data set~\cite{cheng2017remote}, we set the ratios of the number of training set to 10\% and 20\%, respectively, and the rest 90\% and 80\% for testing. All settings are the same as the works in~\cite{cheng2017remote}.

Data augmentation is performed on the training images including original images and attention map, which are randomly rotated by $90^{\circ}$, $180^{\circ}$, $270^{\circ}$, horizontally flipped and vertically flipped. 

Overall accuracy and confusion matrix are two common quantitative evaluation metrics in image classification. The overall accuracy is defined as the number of correctly classified samples, without taking into account the type of category to which they belong, divided by the total number of samples. The confusion matrix is an informative table used to analyze the errors and confusions between different classes, and it gets by counting each class of correct and incorrect classification of the test images and accumulating the results in the table. At the same time, to obtain reliable results on all two data sets,  we repeat the experiment 10 times for each training-test ratio and report the mean and standard deviation of the results.

\subsubsection{Parameter Setting}
Our proposed DDRL-AM method based on CNN models include ResNet-18 and SFT architecture, which are detailedly described in section~\ref{overar}. For our model training, learning rate parameter is one of the important parameters in the whole learning process. In the end-to-end learning phase, The learning rate is set to 0.0001 for finetuned CNN model and 0.001 for SFT model and the classification layers. At the same time, we adopted the learning rate decay strategy. Every ten epochs for SFT model and the classification layers, the learning rate is reduced by $\gamma = 0.1$. For center-loss function, we followed the setting in~\cite{wen2016discriminative} and the parameter optimizer used standard SGD (Stochastic Gradient Descent) with a learning rate of 0.5. The scalar $\lambda = 0.5$ is used for balancing the two loss functions.

\begin{figure*}[!htb]
	\begin{center}
		\frame{\includegraphics[height=0.45\textwidth]{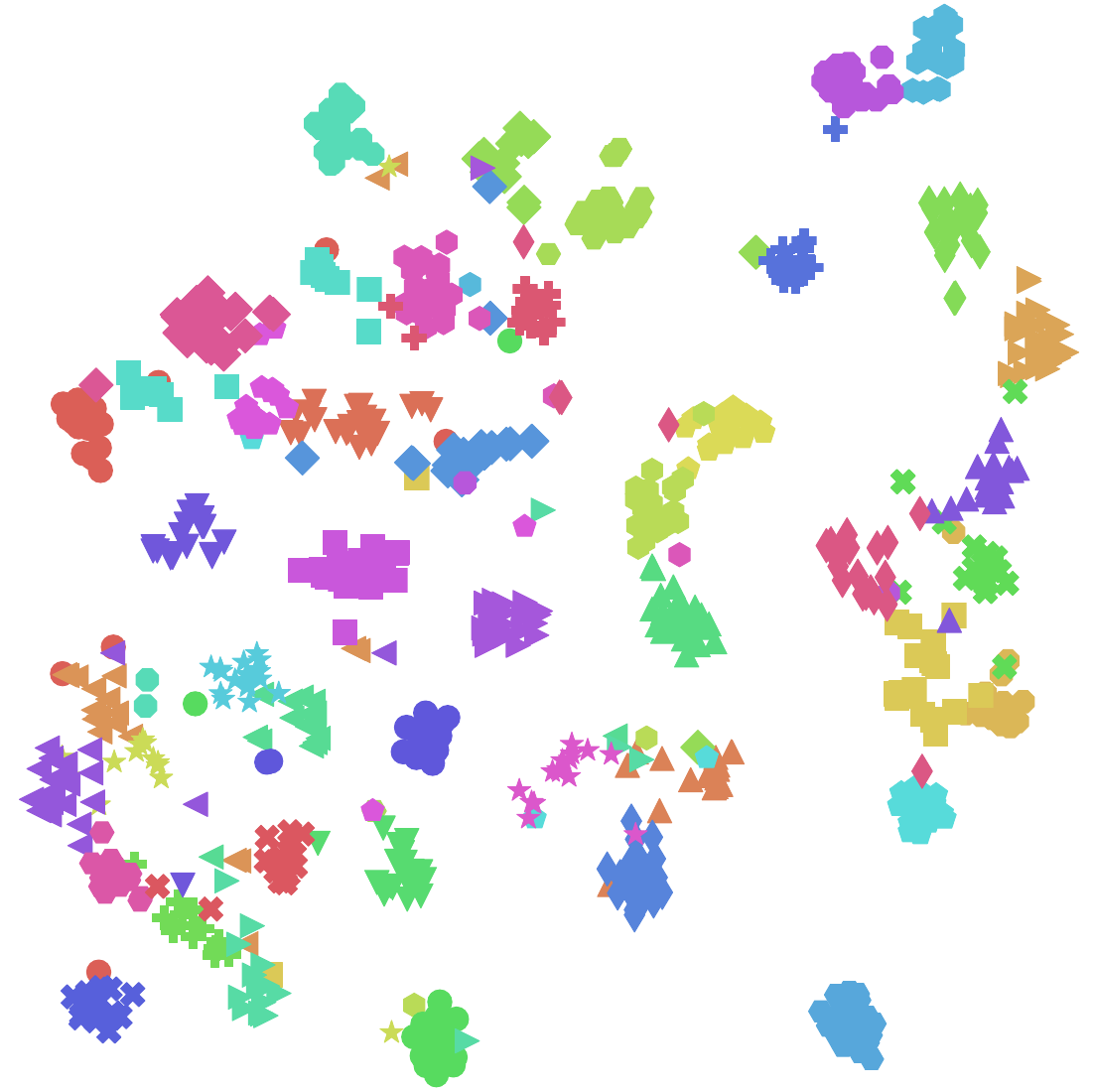}}
		\frame{\includegraphics[height=0.45\textwidth]{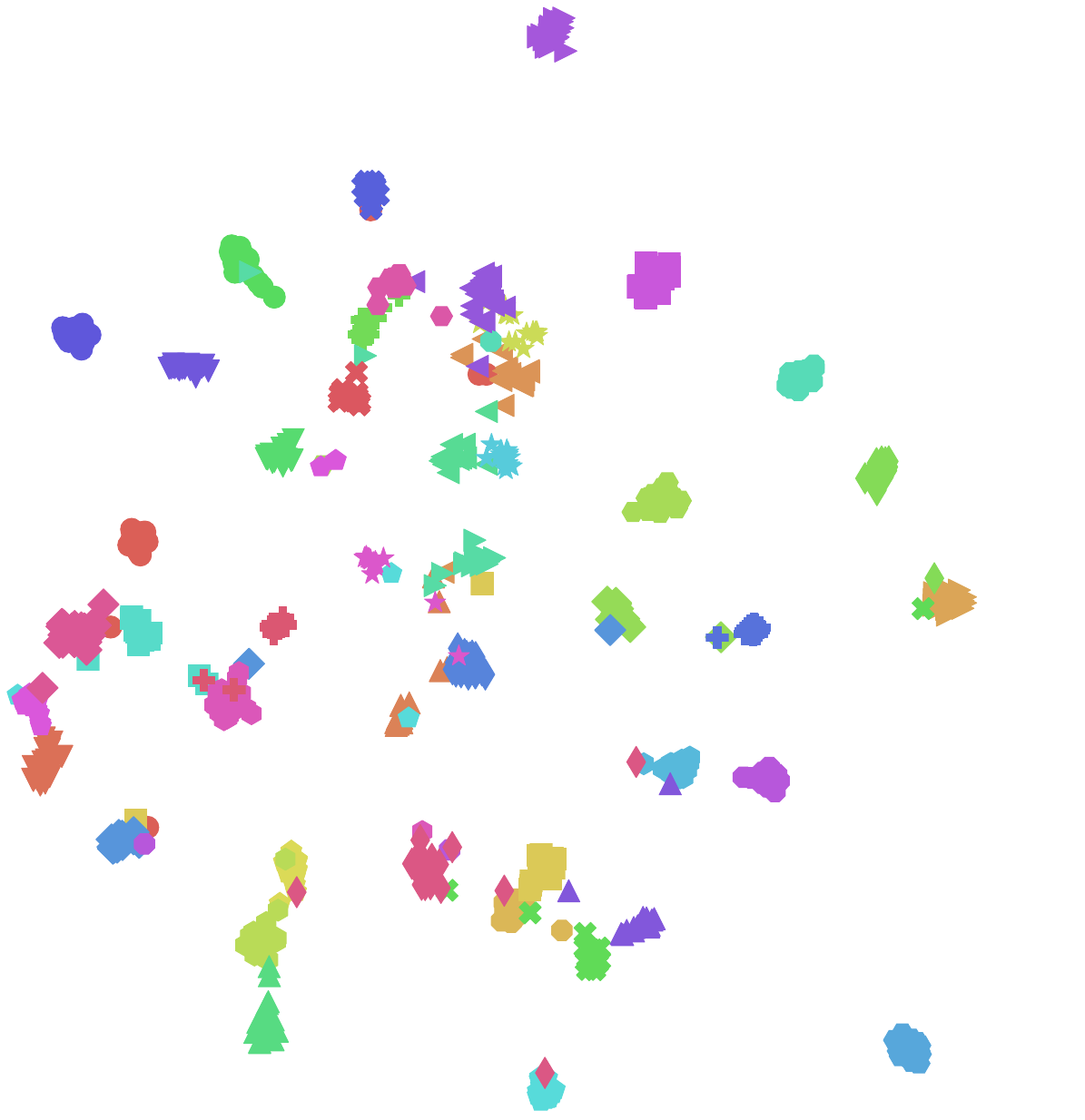}}
		\\ 
		\vspace{1em}
		\includegraphics[width=0.96\textwidth]{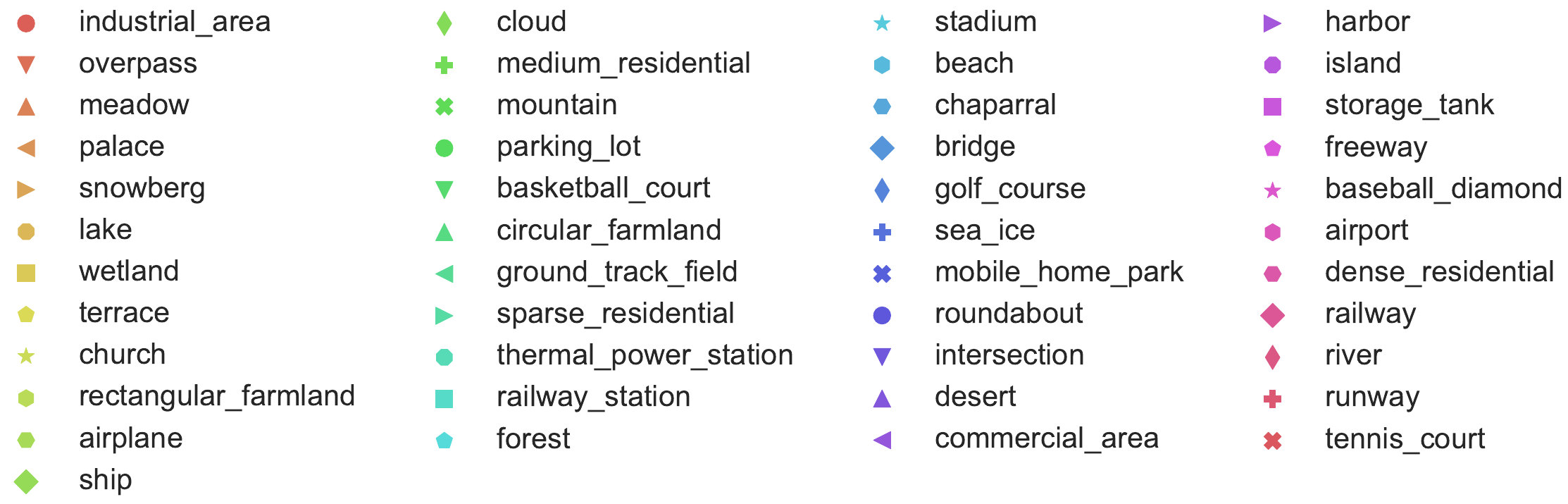}
	\end{center}
	\caption{2-D scatterplots of high-dimensional features generated with t-SNE over the NWPU-RESISC45 dataset. (left) without use center-loss function. (right) use center-loss function. }
	\label{t-snenwpu}
\end{figure*}

\begin{figure*}[!htb]
	\begin{center}
		\frame{\includegraphics[height=0.25\textwidth]{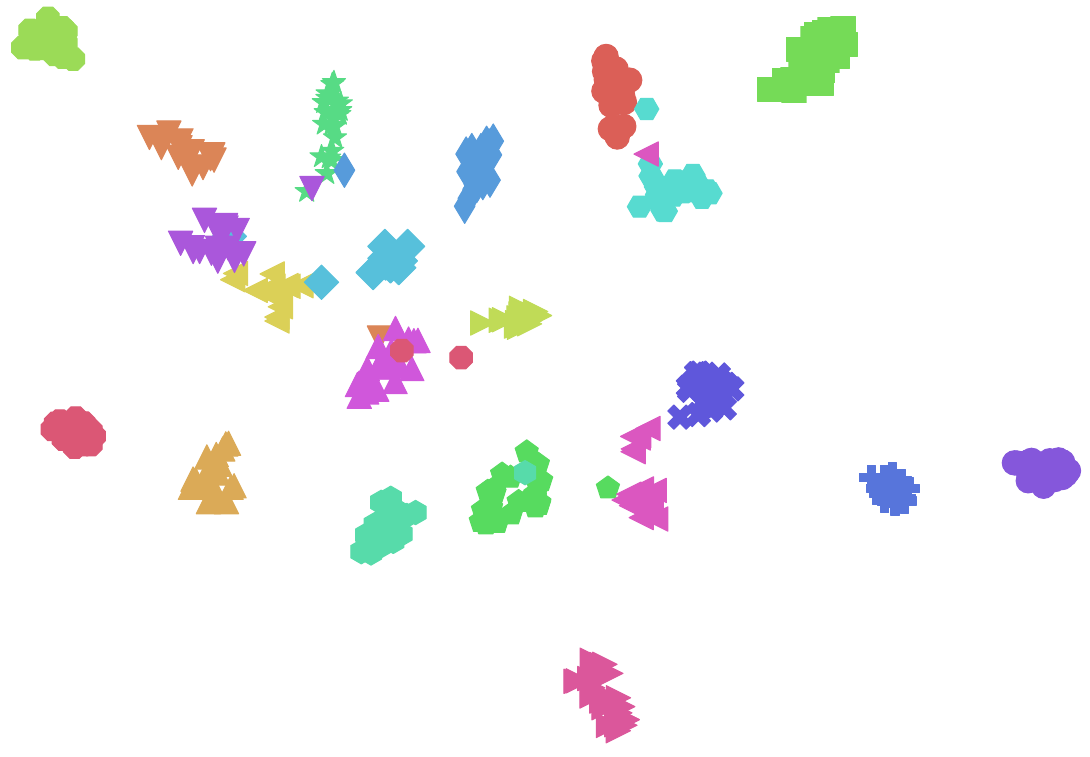}}
		\frame{\includegraphics[height=0.25\textwidth]{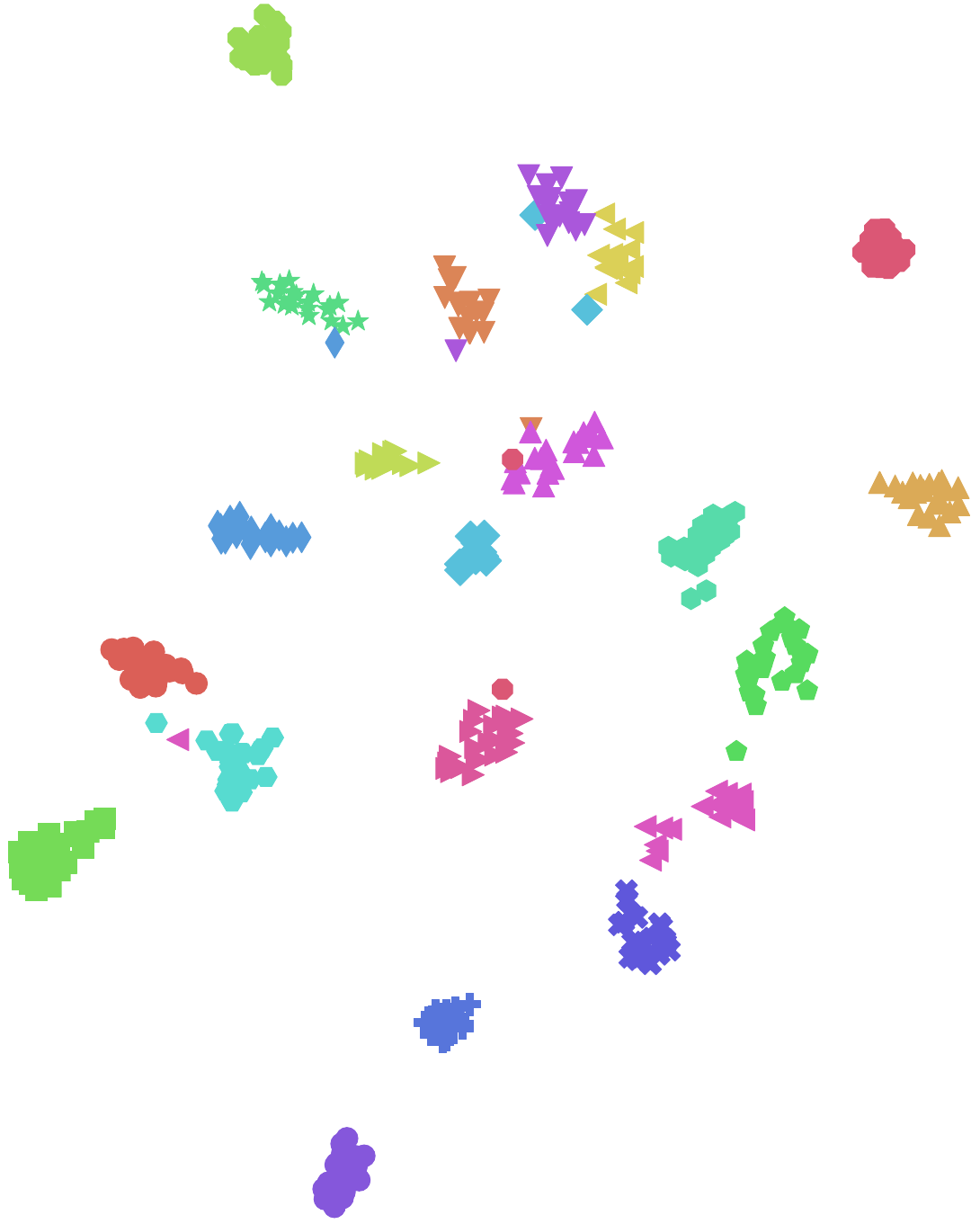}}
		\includegraphics[height=0.25\textwidth]{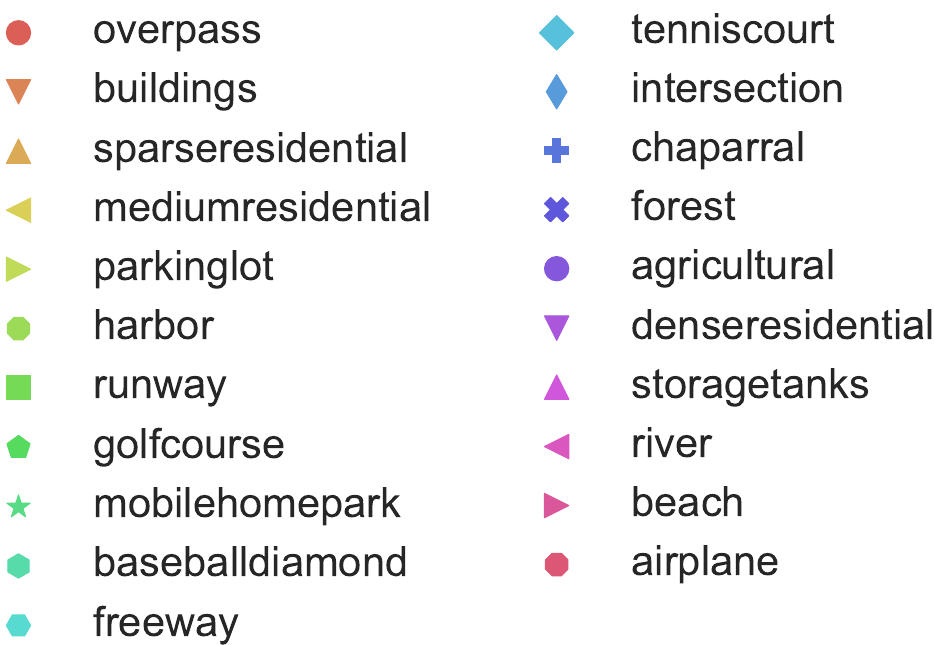}
	\end{center}
	\caption{2-D scatterplots of high-dimensional features generated with t-SNE over the UC Merced dataset. (left) without use center-loss function. (right) use center-loss function. }
	\label{t-sneuc}
\end{figure*}

\begin{figure*}[h]
	\centering
	\includegraphics[width=0.9\textwidth]{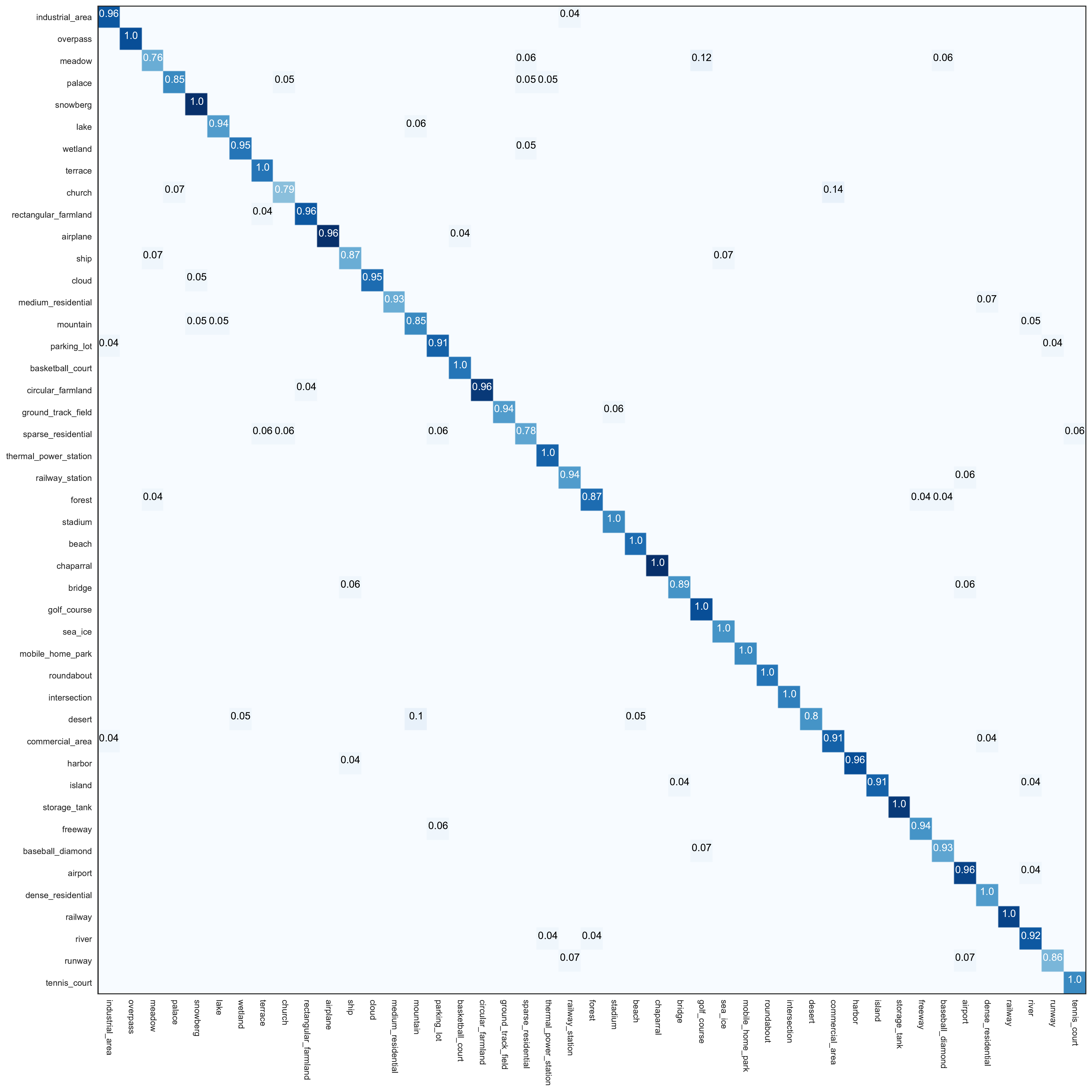}
	\vspace{-0.5em}
	\caption{Confusion matrices of the NWPU-RESISC45 data set under the training ratio of 80\% using DDRL-AM}
	\vspace{-4mm}
	\label{nwpumatrix}
\end{figure*}

\begin{figure*}[!htb]
	\centering
	\includegraphics[width=0.9\textwidth]{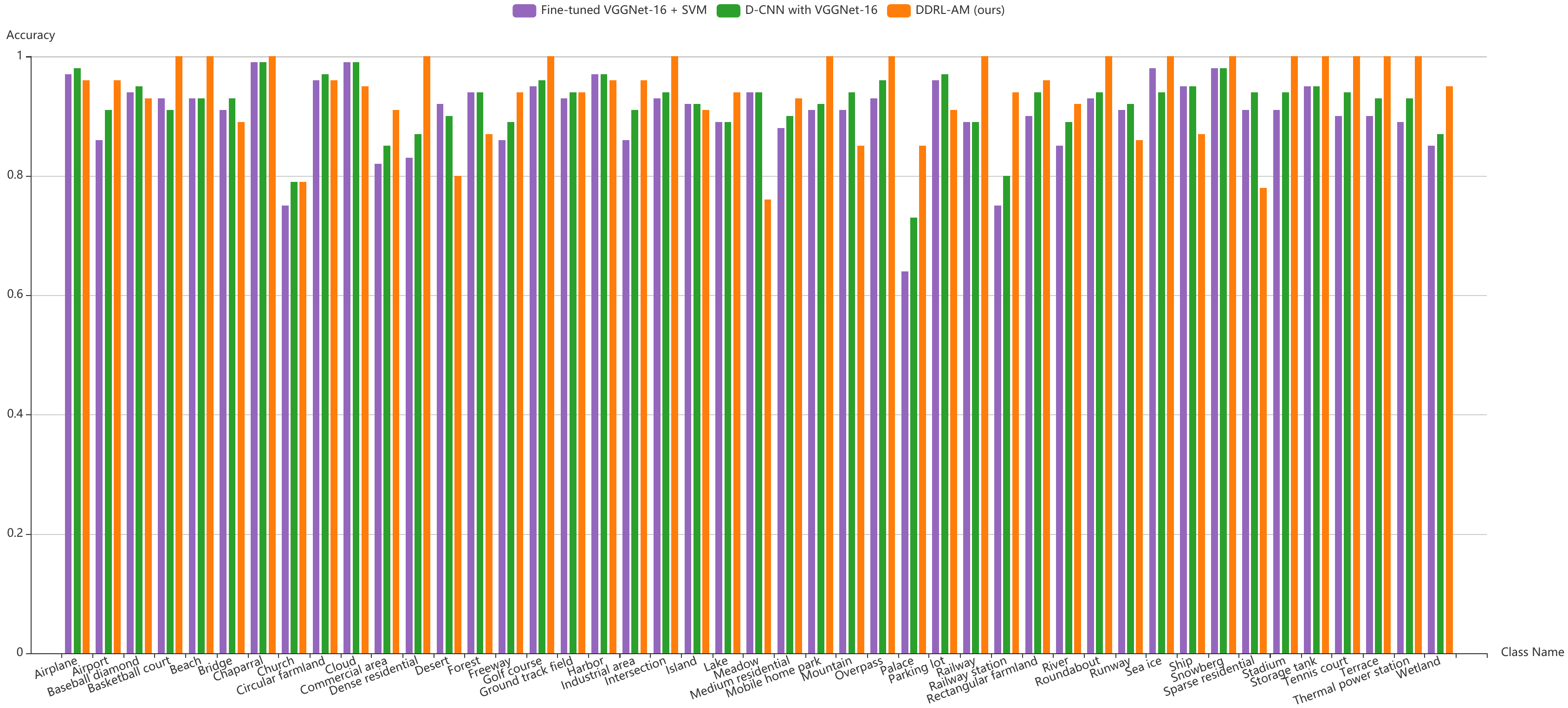}
	\vspace{-0.5em}
	\caption{per-class accuracies of the proposed method and some state-of-the-art references on the NWPU-RESISC45 dataset.}
	\vspace{-4mm}
	\label{nwpucompair}
\end{figure*}

\begin{table*}[t]
	\centering
	\caption{Overall accuracies(\%) of three different setting datasets on different components}
	\scalebox{1}{
		\begin{tabular}{p{3.0cm}<{\centering}  p{3.0cm}   p{2.0cm}<{\centering}}
			\hline
			\textbf{DataSet}&\textbf{Components} &\textbf{Accuracy} \\ \hline
			\multirow{3}{*}{UC Merced} & ResNet-18& $96.19\pm0.14$ \\ 
			&ResNet-18 + AM&$98.24\pm0.10$ \\
			&ResNet-18 + AM+CL&$99.05\pm0.08$ \\ \hline
			\multirow{3}{*}{NWPU-RESISC45-20\%} &ResNet-18& $90.11\pm0.15$\\ 
			&ResNet-18 + AM&$91.03\pm0.13$ \\
			&ResNet-18 + AM+CL&$92.46\pm0.09$ \\ \hline
			\multirow{3}{*}{NWPU-RESISC45-10\%} &ResNet-18& $90.01\pm0.05$\\ 
			&ResNet-18 + AM&$91.84\pm0.05$ \\
			&ResNet-18 + AM+CL&$92.17\pm0.08$ \\\hline
	\end{tabular}}
    \label{diffcom}
	\vspace{-1em}
\end{table*}

\subsection{Experiments and Analysis}
\subsubsection{Generate Attention Map}
The discriminative regions in an image is a set of spatial maps indicating on which area the network focuses. And the attention maps contain information that various layers of the network capture low-, mid-, and high-level representation feature.

Fig.~\ref{atten} illustrates the attention map results from pre-trained ResNet-18 model by Grad-CAM~\cite{selvaraju2017grad} algorithm. From the RGB-saliency mask, we can see that the fine-tuned network learns well to exploit information in target object regions and aggregate features from them. And the attention map essentially shows on where the spatial areas of input the network focus for making output decision. Note that the higher the brightness of the color, the higher the importance of the corresponding area of the image.

\subsubsection{ablation study}
In order to analyze the influence of different components in our framework, we conduct experiments with different setting on two datasets and report the results in Table.~\ref{diffcom}. We use AM to denote the model using the attention mechanism. And CL means the center loss function. By comparing the performances of ResNet-18, ResNet-18+AM, ResNet-18+AM+CL, we find that the two modifications all contribute considerably to the system. We also notice that AM components always leads to a significant improvement.

\subsubsection{Comparisons with other methods}
Since the DDRL-AM use visual attention mechanisms to enrich the power of the CNN feature representations, so we mainly compare our method with CNN feature-based methods including fine-tuned VGGNet-16+SVM~\cite{cheng2017remote} and D-CNNs~\cite{cheng2018deep}.

We report the mean scene classification accuracy (AC) and standard deviation (STD) of the proposed DDRL-AM methods and above mentioned state of the art two methods, on the UC Merced data set~\cite{yang2010bag} and NWPU-RESISC45 data set~\cite{cheng2017remote}, respectively.

Table.~\ref{resultuc} and Table.~\ref{resultnuwp} shows the baseline comparison on two remote sensing scene classification datasets. Our approach provides superior performance compared to existing methods, and our approach provides consistent improvement in performance on most scene categories.

\begin{table}[h]
\centering
\caption{Scene classification results ($ac\% \pm std\%$) using the uc merced data set}
{
\begin{tabular}{llllll}
\hline

\multicolumn{3}{l}{\textbf{Method}} & \multicolumn{3}{c}{\textbf{Accuracy}}  \\ \hline
\multicolumn{3}{l}{BoVW~\cite{yang2010bag}} & \multicolumn{3}{c}{$76.81$}   \\ 
\multicolumn{3}{l}{BoVW + SCK~\cite{yang2010bag}} & \multicolumn{3}{c}{$77.71$}   \\ 
\multicolumn{3}{l}{SIFT + SC~\cite{cheriyadat2014unsupervised}} & \multicolumn{3}{c}{$81.67$}   \\ 
\multicolumn{3}{l}{Unsupervised feature learning~\cite{cheriyadat2014unsupervised}} & \multicolumn{3}{c}{$81.67\pm1.23$}   \\ 
\multicolumn{3}{l}{Fine-tuned GoogLeNet~\cite{castelluccio2015land}} & \multicolumn{3}{c}{$97.10$}   \\ 
\multicolumn{3}{l}{Deep CNN Transfer~\cite{hu2015transferring}} & \multicolumn{3}{c}{$98.49$}   \\ 
\multicolumn{3}{l}{Fusion by addition~\cite{chaib2017deep}} & \multicolumn{3}{c}{$97.42\pm1.79$}   \\      
\multicolumn{3}{l}{Two-Stream Fusion~\cite{yu2018two}} & \multicolumn{3}{c}{$98.02\pm1.03$}          \\
\multicolumn{3}{l}{D-CNN with VGGNet-16~\cite{cheng2018deep}}         & \multicolumn{3}{c}{$98.93\pm0.10$}          \\

\multicolumn{3}{l}{DDRL-AM(ours)}     & \multicolumn{3}{l}{\textbf{$99.05\pm0.08$}} \\ \hline

\end{tabular}}
\label{resultuc}
\end{table}

\begin{table}[h]
	\centering
	\caption{Scene classification results ($ac\% \pm std\%$) using the nwpu-resisc45 data set}
	{
	\begin{tabular}{|cccc}
		\hline
		\multicolumn{2}{c}{\multirow{2}{*}{Method}} & \multicolumn{2}{l}{\textbf{Training ratio}} \\ \cline{3-4} 
		\multicolumn{2}{c}{}                        & 10\%                 & 20\%                 \\ \hline
		\multicolumn{2}{l}{Fine-tuned VGGNet-16~\cite{cheng2017remote}}                      & $87.15\pm0.45$       & $90.36\pm0.18 $      \\
		\multicolumn{2}{l}{D-CNN with VGGNet-16~\cite{cheng2018deep}}    & $89.22\pm0.50$       & $91.89\pm0.22 $      \\ 
		\multicolumn{2}{l}{DDRL-AM (ours)}                    & $92.17\pm0.08$       & $92.46\pm0.09$ \\ \hline
	\end{tabular}}
\label{resultnuwp}
\end{table}

\begin{figure}[t]
	\centering
	\includegraphics[width=0.45\textwidth]{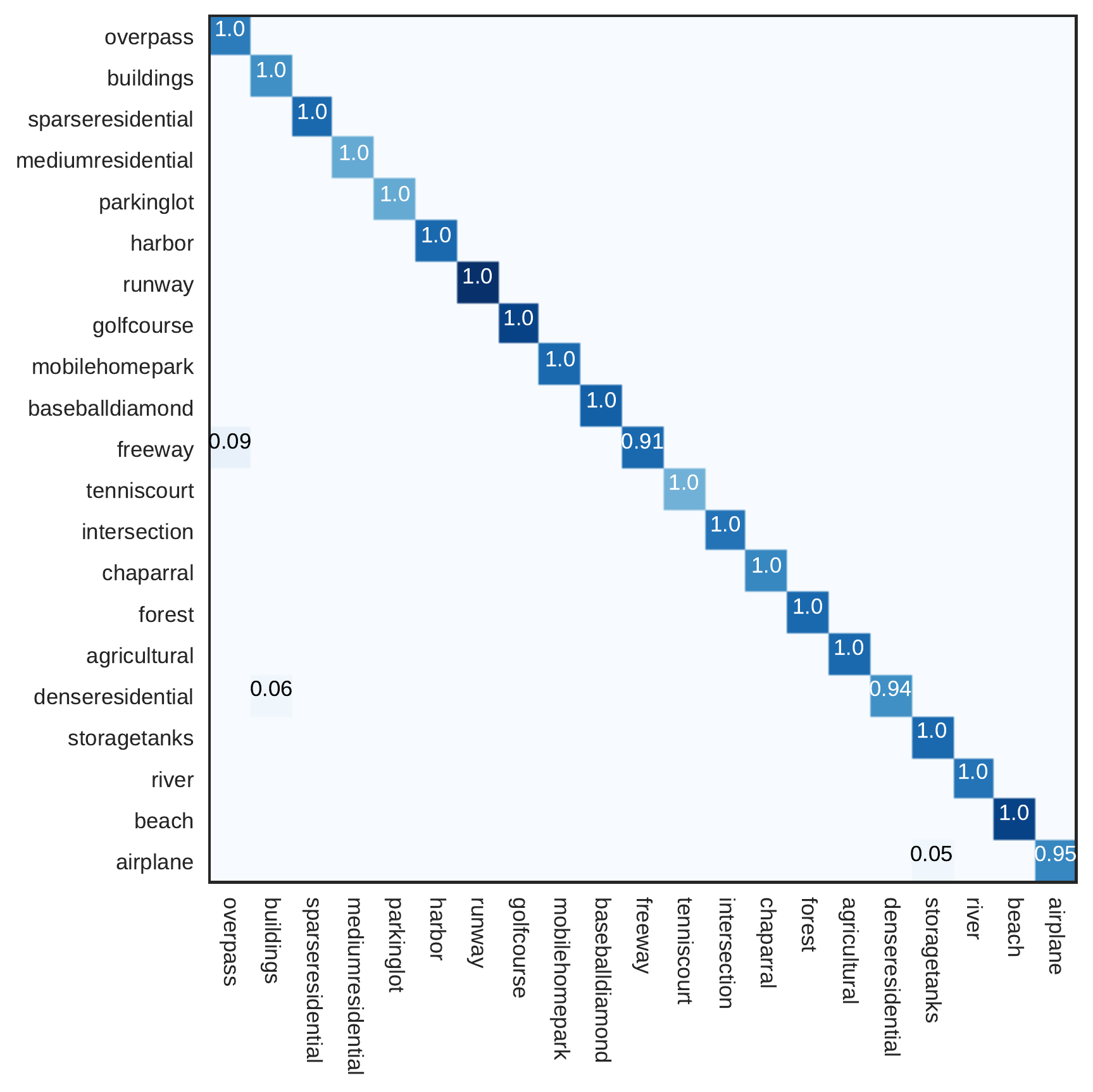}
	\vspace{-0.5em}
	\caption{Confusion matrices of the UC Merced data set under the training ratio of 80\% using DDRL-AM}
	\label{ucmatrix}
\end{figure}
\subsubsection{Statistical Histogram and Confusion Matrices}
In addition to comparing the overall accuracy of the various above algorithms, to further verify distinction performance of our algorithm in different categories, We take a statistical histogram to visually compare the correct rates of various algorithms in different categories. Due to the limitation of space, we report only results under the training ratios of 80\% (for NWPU-RESISC45 data set) and 80\% (for UC Merced data set), For each dataset, we compare the accuracy of the three algorithms mentioned above in each category. Fig.~\ref{nwpucompair} shows the results using fine-tuned VGGNet-16, D-CNN, and our proposed method (DDRL-AM) on the NWPU-RESISC45 dataset. At the same time, in order to deeply analyze the categories of misclassification, we also compute the corresponding confusion matrix, where the entry in the $ith$ row and $jth$ column denotes the rate of test images from the $ith$ class that are classified as the $jth$ class. Fig.~\ref{nwpumatrix} shows the confusion matrices of scene classification results using our proposed method (DDRL-AM) on the NWPU-RESISC45 dataset. Fig.~\ref{ucmatrix} reveals the results on UC-Merced dataset.
From statistical histogram on the NWPU-RESISC45 dataset (Fig.~\ref{nwpucompair}), we can see that our method (DDRL-AM) can be easily distinguished from others, and there are 17 classes obtain the classification accuracy(100\%). Moreover, for palace and railway station categories that are easily misclassified, our method has improved by 0.12 and 0.14 respectively. 

By analyzing the confusion matrix on our method (Fig.~\ref{nwpumatrix}), we can further observe that the number of misclassified categories is relatively reduced and our confusion matrix is quite clean. The most notable confusion is meadow and golf course, which may be caused by the fact that golf course's area is too small and they may both contain a large area of green grass, and thus are easily confused.
From confusion matrix Fig.~\ref{ucmatrix}, we can see the similar phenomenon that the classification accuracy of most classes are close to or even equal to 1. 

By comparing the statistical histogram among the two datasets and analysing confusion matrix, we can conclude that our proposed method has a more powerful discriminating ability, which can make most categories correct to 100

To delve into this results and verify the effectiveness of center loss function, 
We extracted the 25088-dimensional features representations from the first fully connected layer, and we employ the t-SNE~\cite{maaten2008visualizing} algorithm to embed the high-dimensional features in 2-D space for all the scenes of the dataset. We separately visualized two cases, whether or not to use the center-loss target loss function. The respective outcomes can be found in Fig.~\ref{t-snenwpu} and Fig.~\ref{t-sneuc}. By inspecting the derived clusters, it is clear that center-loss function positively affects the abstract semantic information and lead to a large separation in different classes such as overpass and bridge.
More importantly, this function makes all samples with similar semantics closer in high-dimensional space. By exploring the high-dimensional feature space, we can see that our approach can effectively distinguish between different categories.

\section{Discussion}
In this paper, for the three significant challenges of remote sensing image scene classification, namely, within-class diversity, between-class similarity and a small object in scene image, we proposed a discriminative representation method deep discriminative representation learning with attention map (DDRL-AM). From the above experimental results, our approach works well on large scale datasets (NWPU-RESISC45), which can learn better discriminability, reduce visual confusion and boost the performance of remote sensing image scene classification. Our results confirm that the attention map distilled from the pre-trained model has important guiding significance for the classification of the compound image. Although there are important discoveries revealed by these studies, there are also limitations. 
Furthermore, recently, there are many works about attribution methods that depict network has learned by sensitivity maps how different parts of the input contribute to the output. We will try other existing algorithms to generate more meaningful saliency map in the near future.

\section{CONCLUSION}
\label{sec:discussion}
\vspace{-2mm}

In this paper, we have proposed a useful and novel method called DDRL-AM for remote sensing scene classification. And we address the problem of class ambiguity by learning more discriminative feature. Our approach involves the following main tasks: 1) building a two-stream architecture where pre-trained model prior knowledge coded saliency map are used as a second stream and fusing it with the standard RGB stream. And 2) training DDRL-AM that is coupled with a center-loss to obtain discriminative feature for RS images. Extensive experiments are conducted on two benchmark remote sensing scene classification datasets. Our results clearly show that our proposed DDRL-AM methods improve results compared to the current state-of-the-art for remote sensing scene classification.

However, in this paper, we only investigate gradient-based localization method (Grad-CAM) generating the attention map. 
Future work involves investigating alternative saliency image that generates by existing different interpretable algorithms. Another future direction is based on our proposed method to evaluate heat maps highlighting the image regions which bear the most responsibility for the prediction.

\bibliographystyle{IEEEtran}
\bibliography{sceneclassification}

\end{document}